\documentclass[journal]{IEEEtran}
\usepackage{xcolor}
\usepackage[pdftex]{graphicx}
\usepackage{eqparbox}
\usepackage{amsfonts}
\usepackage{multirow}
\usepackage{graphicx}
\usepackage{amsmath}
\usepackage{mathtools}
\usepackage{amssymb}
\usepackage{booktabs}
\usepackage[export]{adjustbox}
\usepackage[pagebackref,breaklinks,colorlinks]{hyperref}
\usepackage{wrapfig}
\DeclareMathOperator*{\argmin}{arg\,min}

\usepackage{hyperref} 
\usepackage{fancyhdr} 

\newcommand{\topnotice}{
    This article has been accepted for publication in \textit{IEEE Transactions on Systems, Man, and Cybernetics: Systems}. 
    This is the author's version which has not been fully edited and \\ content may change prior to final publication.
    Citation information: DOI \href{https://doi.org/10.1109/TSMC.2024.3502498}{10.1109/TSMC.2024.3502498}.
}

\fancypagestyle{everypage}{
    \fancyhf{} 
    \fancyhead[C]{\scriptsize \topnotice} 
    \fancyfoot[C]{%
        \scriptsize
        \textcopyright~2024 IEEE. Personal use is permitted, but republication/redistribution requires IEEE permission. 
        See \href{https://www.ieee.org/publications/rights/index.html}{https://www.ieee.org/publications/rights/index.html} for more information.
    }
}

\begin{document}

\pagestyle{everypage}

\bstctlcite{IEEEexample:BSTcontrol}
    \title{D-LORD for Motion Stylization}
  \author{Meenakshi Gupta,
      Mingyuan Lei,
      Tat-Jen Cham,
      and Hwee Kuan Lee
      \thanks{Meenakshi Gupta, Mingyuan Lei, Tat-Jen Cham  (\texttt{(meenakshi.gupta,mingyuan.lei,astjcham)@ntu.edu.sg}), are associated with Nanyang Technological University, Singapore.} \thanks{Hwee Kuan Lee
  (\texttt{leehk@bii.a-star.edu.sg}) is associated with - Bioinformatics Institute, A$\ast$STAR, Singapore. School of Computing, NUS, Singapore 117417. 
  School of Biological Sciences, NTU, Singapore 639798.
  International Research Laboratory on Artificial Intelligence, Singapore 138632.
  A* Centre for Frontier AI Research (CFAR), Singapore 138632.
  }
  
}

\maketitle
\thispagestyle{everypage}
\begin{abstract}
This paper introduces a novel framework named D-LORD (Double-Latent Optimization for Representation Disentanglement), which is designed for motion stylization (motion style transfer and motion retargeting). The primary objective of this framework is to separate the class and content information from a given motion sequence using a data-driven latent optimization approach. Here, class refers to person-specific style, such as a particular emotion or an individual’s identity, while content relates to the style-agnostic aspect of an action, such as walking or jumping, as universally understood concepts. The key advantage of D-LORD is its ability to perform style transfer without needing paired motion data. Instead, it utilizes class and content labels during the latent optimization process. By disentangling the representation, the framework enables the transformation of one motion sequence’s style to another’s style using Adaptive Instance Normalization. The proposed D-LORD framework is designed with a focus on generalization, allowing it to handle different class and content labels for various applications. Additionally, it can generate diverse motion sequences when specific class and content labels are provided. The framework’s efficacy is demonstrated through experimentation on three datasets: the CMU XIA dataset for motion style transfer, the MHAD dataset, and the RRIS Ability dataset for motion retargeting. Notably, this paper presents the first generalized framework for motion style transfer and motion retargeting, showcasing its potential contributions in this area.
\end{abstract}

\begin{IEEEkeywords}
Motion style transfer, motion retargeting, Double latent optimization for representation disentanglement.
\end{IEEEkeywords}

\section{Introduction}
\IEEEPARstart{M}{otion} stylization is a key technique in animation, computer graphics, virtual reality, games, and robotics. The term ``motion stylization" encompasses both motion style transfer and motion retargeting. Motion style transfer generates animated characters with various emotions from motion capture data. It transfers the style of one motion sequence to another while preserving the original motion content, as shown in Figure \ref{fig:D-LORD}(b). Researchers have employed disentanglement frameworks to separate style from source motion, enabling animators to imbue different emotional expressions or artistic styles into animated characters while maintaining details of the motion \cite{smith2019efficient,park2021diverse}.

The concept of motion style transfer stems from image style transfer \cite{Lee_2018_ECCV}. Previous work, such as Aberman et al.\ \cite{aberman2020}, had applied image-style techniques like disentanglement and Adaptive Instance Normalization (AdaIN) \cite{Huang_2017_ICCV} to motion capture data for emotion transfer. However, they faced challenges in preserving motion content when transferring style between significantly different action types due to using 1D convolution in the content encoder \cite{pang2021image}. Park et al.\ \cite{park2021diverse} addressed this by using a spatial-temporal graph convolution network, improving source motion retention. However, the root trajectory of the generated motion is still calculated directly from the source motion, resulting in an unnatural output when transferring a high-intensity style to a walking motion. They further advanced their work by eliminating the need for labeled motion data and transferring style features locally to different body parts \cite{Lee_2018_ECCV}. This sometimes results in physically implausible whole-body motions due to the independent control of each body part's style. Concurrent work \cite{guo2024generative} has explored generative human motion stylization in the neural latent space, aiming to preserve content while allowing for diverse stylizations. However, it struggles to stylize content motion when the target styles are tied to specific content characteristics.

In summary, existing motion-style transfer algorithms suffer from some of the following drawbacks:
\begin{itemize}
    \item \emph{Adversarial training}: Difficult to train \cite{aberman2020,park2021diverse}.
    \item \emph{Root motion preservation}: Preserving root motion may degrade style transfer quality if the target style is linked to specific content characteristics \cite{aberman2020,park2021diverse,jang2022motion,guo2024generative}, e.g., changing neutral motion to a proud style.
     \item \emph{Spatial relations between joints}: Using 1D convolution for content code generation fails to consider spatial relations between joints, leading to poor motion content preservation for significantly different action types \cite{aberman2020, guo2024generative}, e.g., neutral-kicking and proud-jumping.
    \item \emph{Root trajectory calculation}: Directly calculating the root trajectory from source motion weakens the stylizing effect \cite{aberman2020,park2021diverse,jang2022motion,guo2024generative}.
    \item \emph{Deterministic output}: A pair of input content and style motions yields a deterministic output \cite{aberman2020,jang2022motion}.
\end{itemize}

This paper addresses all these issues and proposes a latent optimization-based disentanglement framework called D-LORD (Double-Latent Optimization for Representation Disentanglement). D-LORD operates in two stages:

Stage 1: Using latent optimization, decompose motion capture data into three latent variables --- class, content, and aleatoric uncertainty (AU) \cite{Kendall_2017_NeurIPS}. 

Stage 2: Train encoders for these latent variables using optimized latent codes, and train a separate variational autoencoder (VAE) \cite{Kingma_2014_ICLR} to map aleatoric uncertainty to a Gaussian distribution space.

D-LORD does not require paired motion data but uses labeled motion data to disentangle class (e.g., emotion, subject ID) and content (e.g., action). Paired data, if available, can further refine the network. It incorporates different class and content labels depending on the specific application. For motion style transfer, emotion is a class label, and action is a content label.

Additionally, D-LORD can be used for motion retargeting, applying a source motion to a target character with different kinematic properties \cite{villegas2018neural,nikpour2023spatial}. The skeletal differences between the source and target characters necessitate disentangling skeleton-independent features of the source motion and transferring them to the target character \cite{lim2019pmnet,aberman2020skeleton}. D-LORD effectively performs motion retargeting by treating the class label as a person's identity and the content label as the action.

Disentanglement is crucial for both motion style transfer and motion retargeting. Current algorithms typically focus on style-dependent features for motion style transfer or style-independent features for motion retargeting, lacking a comprehensive solution for both. In motion style transfer algorithms, the source motion is applied to the target after passing through the instance normalization layer, with the root trajectory of the target motion being directly calculated from the source motion. Conversely, motion retargeting algorithms directly use skeletal information, provided in the form of a T-pose. Thus, current motion style transfer algorithms cannot be used for motion retargeting, and vice versa.
The D-LORD framework addresses this by disentangling both feature types, making it applicable to both tasks based on selected class and content labels.

The contributions of our work are:
\begin{itemize}
    \item \emph{Ease of training}: Our framework uses latent optimization, simplifying the training process without requiring adversarial training.
    \item \emph{Accurate motion style adaptation}: By disentangling class and content features, our algorithm adapts style-specific motion characteristics to the content, even when the target style is linked to specific motion characteristics.
    \item \emph{Style transfer between different actions}: The content encoder is trained to generate optimized embeddings, allowing style transfer between significantly different action types despite using 1D convolution.
     \item \emph{No strong motion content preservation}: Our framework does not rely on calculating the root trajectory from the source motion, avoiding issues with strong motion content preservation.
    \item \emph{Generalized disentanglement}: It incorporates various class and content labels, applicable to both motion style transfer and motion retargeting.
    \item \emph{Diverse motion data generation}: It can generate diverse motion capture data by sampling AU latent from a Gaussian distribution, enhancing animation variety and richness.
\end{itemize}

In summary, D-LORD is a generalized motion stylization framework that is easy to train. It avoids strong motion content preservation, allowing for effective style transfer between different action types. Additionally, it generates diverse motion sequences. Through various experiments, we demonstrate the ability of our method to produce improved results in terms of visual quality, style transfer, and content preservation.

\section{Related work}\label{section:LS}
\subsection{Image Style Transfer} 
Early methods utilized deep features from a pre-trained deep convolutional network by the Visual Geometry Group (VGG) \cite{simonyan2014very} for image classification, extracting style features from shallower layers and content features from deeper layers. However, these methods were computationally inefficient due to iterative optimization. Johnson et al.\ \cite{johnson2016} introduced perceptual loss functions and feed-forward generator networks for efficient, single-pass stylized image generation. Ulyanov et al.\ \cite{ulyanov2017instance} proposed instance normalization (IN) to remove instance-specific contrast from content images and later extended it to AdaIN \cite{Huang_2017_ICCV}, which aligns the mean and variance of content features with style features.

Generative adversarial networks (GANs) \cite{creswell2018generative} were used to align translated image distributions with real images in the target domain \cite{Lee_2018_ECCV, Huang_2018_ECCV}. Lee et al.\ \cite{Lee_2018_ECCV} proposed disentangling content and domain-specific attributes for diverse outputs, while others decomposed images into domain-invariant content and style codes \cite{Huang_2018_ECCV}. These methods were limited to two domains, but Choi et al.\ \cite{choi2018stargan1, choi2020stargan2} introduced Star Generative Adversarial Network (StarGAN) for multi-domain translations using a single model. Despite the advances in disentanglement, GANs are difficult to train and require careful tuning.

To address training challenges, Gabbay and Hoshen introduced latent optimization for representation disentanglement (LORD), a non-adversarial approach to image style transfer \cite{Gabbay2020}. In their later work \cite{gabbay2021image}, they proposed a method to disentangle partially labeled factors and separate residual factors (unknown factors of variation). The aleatoric uncertainty in motion datasets can be defined similarly to residual factors in their paper. However, this approach requires replacing the generator with StyleGAN2 and adding an adversarial discriminator for real images. For more on image style transfer, see \cite{pang2021image, jing2019neural, alotaibi2020deep}.

\subsection{Motion Style Transfer}
Early motion style transfer methods \cite{wu2006line, xia2015realtime} used statistical properties of joint quaternions and local mixtures of autoregressive models to define motion style. These approaches required paired datasets and preprocessing, limiting their scalability and applications.

Inspired by image style transfer, research shifted to data-driven frameworks for motion style transfer. Holden et al.\ \cite{holden2016deep} extended Gatys et al.'s work \cite{gatys2016} for human motions by training an autoencoder to match hidden unit activations and Gram matrices. They later replaced optimization with a feed-forward network \cite{holden2017fast}, but content and style were not disentangled. The method in \cite{du2019stylistic} used a conditional VAE to learn the style distribution and disentangle these features. Dong et al.\ \cite{dong2020adult2child} applied a cycle-consistent adversarial network (CCycleGAN) \cite{Zhu_2017_ICCV} to transform adult to child motions, though it struggled with non-cyclic motions. In \cite{wang2021cyclic}, content and style features were combined using a pre-trained network, CCycleGAN, and kinematic constraints for improved motion style transfer.

Recent deep-learning frameworks \cite{aberman2020, park2021diverse, jang2022moti} have utilized disentanglement and AdaIn from image style transfer. Aberman et al.\ \cite{aberman2020} encoded motions into content and style codes but struggled with significantly different actions due to 1D convolution in the content encoder \cite{pang2021image}. Park et al.\ \cite{pang2021image} improved upon this by using a spatial-temporal graph convolution network but faced unnatural outputs when transferring high-intensity styles to walking motions. They later improved their work \cite{jang2022motion} by removing the need for labeled data. Jian et al.\ \cite{pan2021fast} proposed a fast human motion transfer method based on a meta-network \cite{shen2018neural}, though the results contained noise. Tao et al. \cite{tao2022} supported online motion style transfer but required manual calibration of bones and phases. Concurrent work \cite{guo2024generative} generates diverse motion stylization in the neural latent space but struggles to stylize content motion when target styles are tied to specific content characteristics.

\subsection{Motion Retargeting} 
Traditional motion retargeting methods \cite{gleicher1998retargetting, lee1999hierarchical, tak2005physically, choi2000online} relied on Inverse Kinematics (IK) and iterative optimization with manually designed kinematic constraints, often resulting in unnatural motions.  Recent approaches use deep neural networks for more realistic retargeting.

Jang et al.\ \cite{jang2018variational} developed a deep learning-based system using a deep VAE, combining a deep convolutional inverse graphics network (DC-IGN) and a U-Net. This approach required paired training data, limiting its usability. Villegas et al.\ \cite{villegas2018neural} introduced an unsupervised motion retargeting algorithm with two recurrent neural networks (RNNs) and a forward kinematics layer to discover joint rotations without ground-truth data. Lim et al.\ \cite{lim2019pmnet} improved this with their pose-movement network (PMnet), which disentangles frame-by-frame poses and overall movement for better fitting to target characters. However, these data-driven methods rely on adversarial training, making them challenging to train.

Our D-LORD framework uses latent optimization for non-adversarial disentanglement, separating skeletal structure (identity) and motion for effective motion retargeting.

\section{D-LORD framework for motion stylization} \label{section:method}

\begin{figure*}
  \centering
  \includegraphics[width=0.9\textwidth]{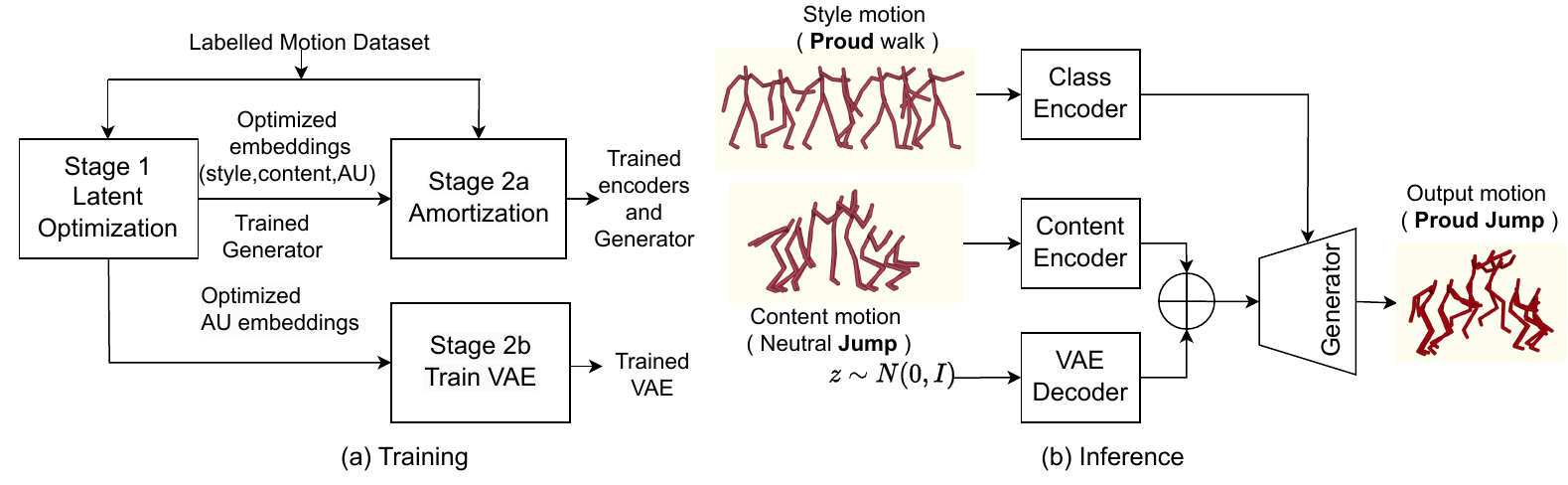}
   \caption{D-LORD framework for Motion stylization.}
   \label{fig:D-LORD}
\end{figure*}

This section presents our D-LORD framework for motion stylization, inspired by image-style transfer works \cite{Gabbay2020, Huang_2017_ICCV, Huang_2018_ECCV, karras2019style}.
 D-LORD builds upon LORD, a latent optimization framework proposed by Gabbay and Hoshen \cite{Gabbay2020}. LORD is a class-supervised image disentanglement method that utilizes shared latent optimization, asymmetric regularization, and a second amortization stage for single-shot generalization. Although LORD, combined with bone length consistency loss, can perform motion style transfer and retargeting, it faces common motion style transfer issues, such as preserving root motion. When adapted for motion retargeting, LORD cannot handle advanced tasks like transferring personalized skeleton features (e.g., gait cycle) or generating diverse motion sequences for a given target skeleton.
 
Unlike images, human motion data is more intricate due to temporal variations and a different latent manifold structure. Thus the architecture of LORD, designed for image-style transfer, must be adapted to handle aleatoric uncertainty \cite{Kendall_2017_NeurIPS} in human motion, even in repeated tasks by the same individual. To address this, we developed D-LORD, which disentangles motion capture data into three latent spaces. In addition to class labels used in LORD, D-LORD uses content labels for latent optimization and adds a skeleton consistency loss to encourage limb lengths to remain invariant in the synthesized motion sequence. To generate diverse motion sequences, the aleatoric uncertainty latent is mapped to a Gaussian distribution using a VAE in the second stage.

This section first outlines the motion representation used in this framework and then details the methodology.

\subsection{Motion Description and Motion Representation}
\emph{Motion description:}
A motion sequence can be described as consisting of three parts: class information, content information, and aleatoric uncertainty. Consider the case of the Berkeley Multimodal Human Action Database (MHAD), which contains 11 actions performed by 12 subjects, with each action repeated five times.
\begin{itemize}
    \item \emph{Class information}. This represents the subject-dependent aspect of the motion and is indicated by the class label. In the MHAD dataset, it corresponds to the subject's identity, which remains the same across all frames and can be used to categorize or identify individual subjects.
    
    \item \emph{Content information}. This represents the subject-invariant aspect of the motion that is consistent across the population and is indicated by the content label. In the MHAD dataset, it refers to the type of action being performed.
    \item \emph{Aleatoric uncertainty (AU)}. It captures the variations between repetitions of the same action by the same subject, accounting for unpredictable factors, such as variations in movement, speed, or style.
\end{itemize}
Motion sequences can be effectively described using these three components: class, content, and aleatoric uncertainty. This representation is valuable for applications like action recognition, motion synthesis, motion stylization, and understanding human behavior.

\emph{Motion representation:}
In our setting, we represent motion sequences using trajectories of body markers in a 3D space. Let $J$ be the number of body markers used to represent a human skeleton in a given frame, with each body marker represented by 3D position coordinates $(x,y,z)$. Thus, $j_{i}^{t, k} \,{\in}\, \mathbb{R}^3 $ denotes the $(x, y, z)$ coordinates of the $k^{\text{th}}$ marker at frame $t$ of $i^{\text{th}}$ motion sequence. The $t^{\text{th}}$ frame of the $i^{\text{th}}$ motion sequence $m_i^t$ can be described as $m_i^t \,{=}\, (j_i^{t,1},\dots,j_i^{t,J}) \,{\in}\, \mathbb{R}^{3 \times J}$. A motion sequence $M_i$ is an ordered sequence of $T$ frames, represented as $M_i \,{=}\, (m_i^1,\dots,m_i^T) \,{\in}\, \mathbb{R}^{3 \times J \times T}$. To make all motion sequences comparable, they are down-sampled or up-sampled to have the same length of $T$ frames. To ensure a common starting point for all motion sequences, first-pose root normalization is performed by subtracting the root position of the first frame from all markers in all frames. Orientation normalization is also performed to unify the direction in which the subject faces at the beginning of the motion. This is achieved by rotating the z-axis so that the subject faces the positive x-axis, with the pelvis parallel to the y-axis. Finally, all motion sequences are normalized to have coordinate values between 0 and 1.

\subsection{Double-Latent Optimization for Representation Disentanglement (D-LORD) Methodology}
D-LORD disentangles human motion sequences using a two-stage process, as shown in Figure \ref{fig:D-LORD}. In Stage 1, the algorithm optimizes embeddings for class, content, and aleatoric uncertainty (AU) using latent optimization. The term ``double latent optimization'' refers to the optimization of both class and content embeddings.  In Stage 2, encoders are trained to estimate the optimized embeddings obtained from Stage 1 for new input motion sequences. Given an input motion sequence, the encoders produce the corresponding class, content, and AU embeddings. Additionally, a separate VAE network is trained to map the aleatoric uncertainty latent to another latent space and align it with a prior Gaussian distribution using Kullback-Leibler (KL) divergence. During inference, the VAE allows sampling of the AU latent from the Gaussian distribution, enabling the generation of diverse human motion sequences, even for a single subject performing the same action.

\subsubsection{Stage 1 -- latent optimization}
\begin{figure}
  \centering
  \includegraphics[width=0.9\columnwidth]{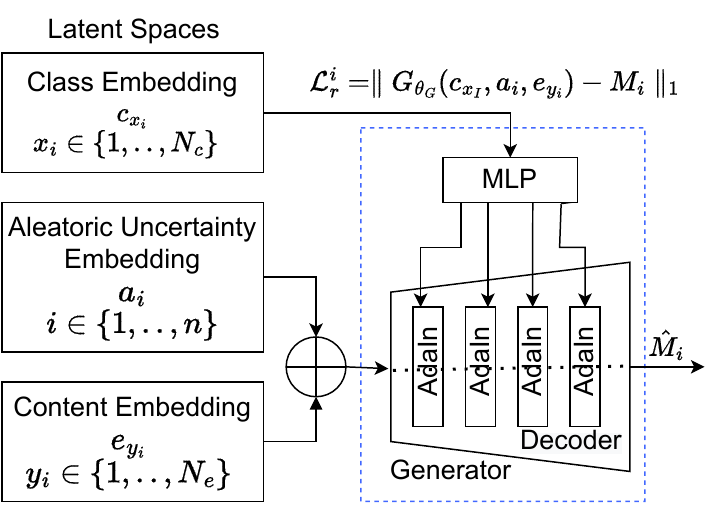}
   \caption{Stage 1: Latent optimization. Parameters of the generator and all the class, content, and AU embeddings are optimized using SGD optimizer. All motion sequences of the same class share a single class embedding and those of the same content share a single content embedding. The network is trained using Equation \ref{eq_4}. Once the network is trained, the latent spaces of the training set are disentangled.}
   \label{fig:stage1}
\end{figure}

Figure \ref{fig:stage1} illustrates the diagram for this stage. Our approach incorporates two labels (class and content) and employs three types of embeddings (class, content, and AU) for the motion sequences. 

Consider a dataset comprising $n$ motion sequences, denoted as  $M_1, M_2, \dots, M_n \,{\in}\, \mathcal{M}$. Let $N_{c}$ be the number of unique class labels (e.g., subject IDs) and $N_{e}$ the number of unique content labels (e.g., action types) for the given dataset. For each motion sequence $M_i$, we are provided with a class label $x_i \,{\in}\, \{1, \dots, N_{c} \}$ and a content label $y_i \,{\in}\, \{1, \dots, N_e \}$. We represent the class embedding of a given class label $x_i$ as $c_{x_i}$ and the content embedding of a given content label $y_i$ as $e_{y_i}$. There are $N_{c}$ class embeddings, each corresponding to a unique class label, and $N_e$ content embeddings, each corresponding to a unique content label. All the motion sequences with the same class label share a common class embedding, while all the motion sequences with the same content label share a common content embedding. The aleatoric uncertainty (AU) embedding, denoted as $a_i$, captures the variations that differentiate motion sequences with identical class and content labels. Each motion sequence has a unique AU embedding, resulting in a total of $n$ AU embeddings for the dataset.

We assume that the motion sequences can be disentangled into representations in three latent spaces, $\mathcal{X}$, $\mathcal{Y}$, and $\mathcal{A}$. The primary objective of this stage is to disentangle the class embedding $c_{x_i} \,{\in}\, \mathcal{X}$, the content embedding $e_{y_i} \,{\in}\, \mathcal{Y}$, and the AU embedding $a_i \,{\in}\, \mathcal{A}$ for a given motion sequence $M_i$. This disentanglement allows the generator $G$, which consists of a multi-layer perceptron (MLP) and a decoder network, to transform these disentangled embeddings into a motion sequence. Thus, the reconstruction loss between the generated motion sequence $\hat{M}_i$ and the actual motion sequence $M_i$ can be expressed as:
\begin{equation}\label{eq_1}
    \mathcal{L}_r^i =  \lVert G_{\theta_G}(c_{x_i},a_i,e_{y_i}) - M_i \rVert_2 
\end{equation}
where $\theta_G$ represents the parameters of the generator network. 

In this stage, we optimize the class and content embeddings directly using latent optimization. As the class embedding is shared exactly among all motion sequences with the same class label, it is impossible to include any content information in the class embedding. Similarly, as the content embedding is common across all motion sequences with the same content label, it is also impossible to include any class information in the content embedding. We learn the AU representation by optimizing over per-sample AU embeddings using latent optimization. 

In the proposed network architecture, the class embedding is incorporated into the AdaIN layers of the generator using an MLP network. The AdaIN layer applies an affine transformation to the feature activations, modifying the mean and variance of the channels. Since this affine transformation is temporally invariant and only affects non-temporal attributes of the motion, the class label used for class embeddings should correspond to the time-invariant properties of the motion sequence. In contrast, the content embedding is provided to the input layer of the generator and is responsible for motion generation. Therefore, the content label is associated with the deterministic time-varying aspects of the motion sequence. The AU embedding captures random variations in the motion sequences and is given as input to the generator along with the content embedding. In the proposed network architecture, the AU embedding is designed to be independent of the class label, but it may still capture some content-related information if there are significant intra-content variations. To minimize the leakage of content information into the AU embedding, the AU embedding is regularized to minimize information. This regularization is achieved by initializing $a_i$ from Gaussian noise with a random mean $\epsilon_i$ and a fixed variance $\sigma^2$, i.e.\ $a_i \,{\sim}\, \mathcal{N}(\epsilon_i,\sigma^2I)$, and adding an $L_2$ regularization term for the AU embedding in the objective function, as done in \cite{Gabbay2020}.

To ensure consistent bone lengths in the synthesized motion sequence $\hat{M}_i$ over time, a skeleton-consistency loss is introduced into the optimization function \cite{mofusion}. The skeleton-consistency loss aims to minimize the temporal variance of bone lengths and is calculated as follows:
\begin{equation}\label{eq_2}
    \mathcal{L}_s^i =  \frac{1}{T-1} \sum\limits_{t=1}^T (l_t^i - \bar{l^i})^2
\end{equation}
where $\bar{l^i}$ is the vector of mean bone lengths of the generated motion sequence $\hat{M}_i$ and $l_t^i$ is the vector of bone lengths at frame $t$ of the generated motion sequence $\hat{M}_i$. Thus, the final objective function of Stage 1 becomes:
\begin{equation}\label{eq_3}
    \mathcal{L}_1 =  \sum_{i=1}^n \mathcal{L}_r^i + \lambda \lVert a_i \rVert_2 + \mathcal{L}_s^i 
\end{equation}
where $\lambda$ is a constant for $L_2$ regularization. The first term in the overall Stage 1 objective function is the reconstruction loss $\mathcal{L}_r^i$, which measures the discrepancy between the synthesized motion sequence $\hat{M_i}$ and the input motion sequence $M_i$. It encourages the generator to produce motion sequences that resemble the original input sequences. The second term is $L_2$ regularization ($\lambda \lVert a_i \rVert_2$), which is applied to the AU embedding $a_i$ to discourage the embedding from carrying excessive information by penalizing its magnitude. The third term is the skeleton-consistency loss ($\mathcal{L}_s^i$), which ensures that the bone lengths in the synthesized motion sequence remain consistent across time. Once the model is trained in Stage 1, all motion sequences are disentangled into three optimized latent embeddings: class, content, and AU. Stage 1 is followed by Stage 2a, where we train the encoders to estimate the optimized embeddings learned in Stage 1 for new input motion sequences.

All latent codes and the parameters of the generator are learned end-to-end using stochastic gradient descent:
\begin{gather}
    \mathbf{c}^*, \mathbf{a}^*, \mathbf{e}^*, \theta_G^* = \argmin_{\mathbf{c}, \mathbf{a}, \mathbf{e}, \theta_G} \mathcal{L}_1 \notag\\ 
    \mathbf{c} = (c_1, \dots, c_{N_{c}}),\>
    \mathbf{a} = (a_1, \dots, a_n),\>
    \mathbf{e} = (e_1, \dots, e_{N_e})
    \label{eq_4}
\end{gather}

\subsubsection{Stage 2a -- Training encoders for amortization inference}

\begin{figure}
  \centering
  \includegraphics[width=0.9\columnwidth]{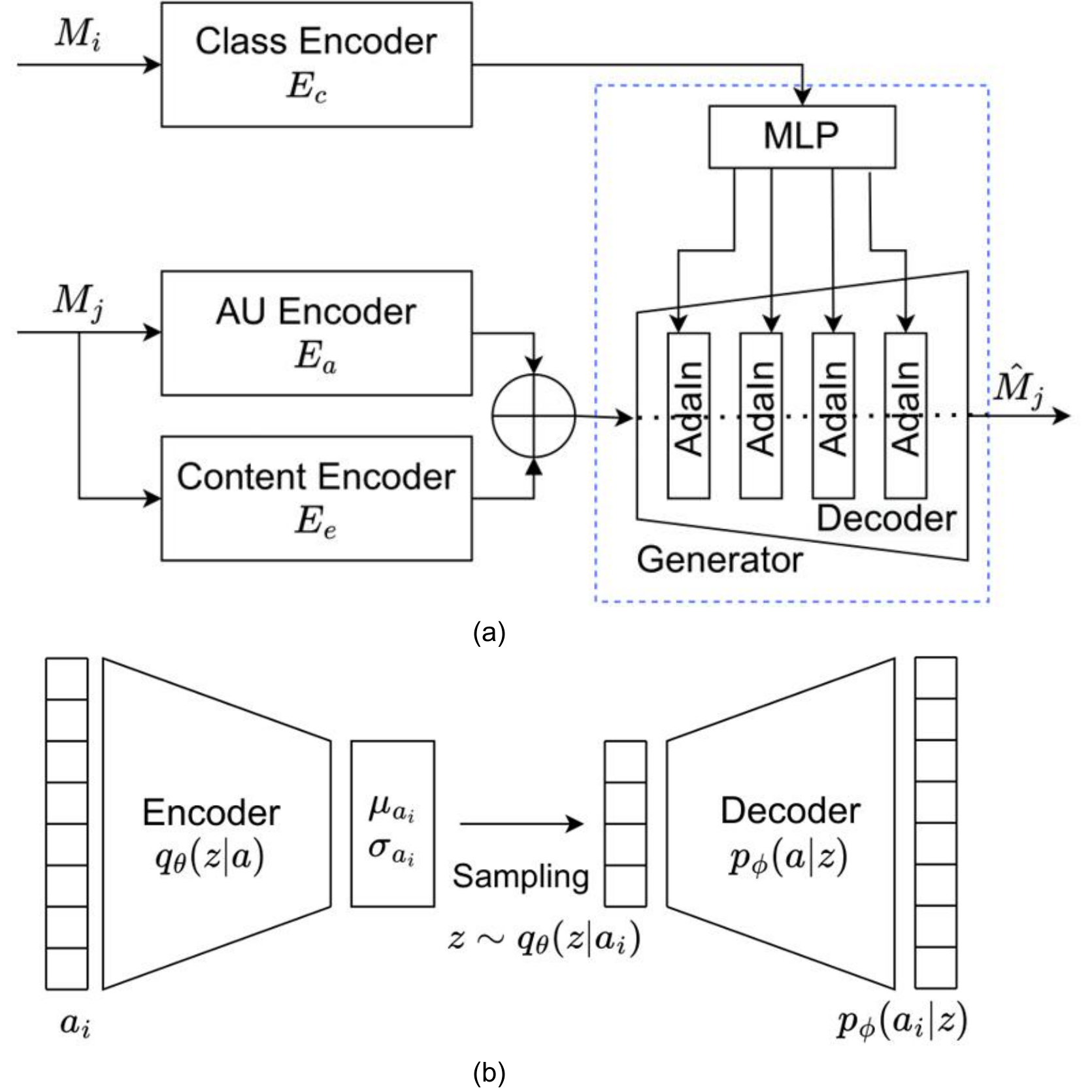}
   \caption{Stage 2: (a) Amortization. All three encoders are trained to generate the optimized embeddings of stage 1 for a given motion sequence. The model is trained using Equation \ref{eq_6}. (b) Variational autoencoder that is used to map AU latents to another latent space that follows the Gaussian distribution.}
   \label{fig:stage2}
\end{figure}

In Stage 1, we optimized the latent variables using Equation \ref{eq_4}. Stage 1 can generate a motion sequence given the motion sequence labels (optimized embeddings corresponding to labels). However, given a new motion sequence in which the labels are unknown, the embeddings need to be estimated. Our objective for this framework is to disentangle the class and content information from a given motion sequence. Thus, in this stage, we train three encoders: Class, Content, and AU, to predict the optimized embeddings learned in Stage 1 for a given motion sequence. Figure \ref{fig:stage2}(a) illustrates the architecture, where each encoder outputs its respective embedding. The objective function aims to minimize the error between the embeddings estimated by the encoders and the original embeddings learned in Stage 1. A reconstruction loss term is added to the objective function to ensure that the learned embeddings accurately reconstruct the original motion sequence. Similar to Stage 1, a skeleton consistency loss is included in the optimization function to maintain consistent bone lengths in the synthesized motion sequence over time. Thus, the overall loss function for Stage 2a becomes: 
\begin{align}\label{eq_5}
  \mathcal{L}_2 =  \sum_{j\in n,{i=1}}^n& \lVert G_{\theta_G}(E_{c}(M_i), E_{a}(M_j),E_e(M_j) - M_j \rVert_2 
  \notag\\
      &+ \alpha_1 . \lVert (E_{c}(M_i)- c_{x_i}) \rVert_2
  \notag\\
     &+  \alpha_2 . \lVert (E_{a}(M_j)- a_j) \rVert_2
  \notag\\
       &+  \alpha_3 . \lVert (E_e(M_j)- e_{y_j}) \rVert_2 +  \mathcal{L}_s^i  
\end{align}
The objective function consists of several terms: the reconstruction / cross-construction loss, three regularization terms that encourage the encoders to match the learned embeddings from Stage 1, and the skeleton consistency loss. The coefficients $\alpha_1, \alpha_2$, and $\alpha_3$ control the importance of these regularization terms. The training procedure depends on whether paired motion sequences are available in the training dataset. If paired sequences are unavailable, the network is trained only with the reconstruction loss ($M_j \,{=}\, M_i$). However, if paired sequences are available, the network is trained alternately for reconstruction and cross-construction loss. Cross-construction loss involves using two different motion sequences ($M_j \,{\neq}\, M_i$) that share the same class labels but have different content labels. The parameters of the encoders $E_{c}$, $E_{a}$, and $E_e$ are randomly initialized, while the parameters of the MLP and the decoder network are initialized from Stage 1, leveraging the knowledge obtained from the previous stage.

The parameters of all three encoders ($E_{c}, E_{a},$ and $E_e$), the MLP network, and the decoder are learned end-to-end using stochastic gradient descent:
\begin{equation}\label{eq_6}
\theta_{c}^*, \theta_{a}^*, \theta_{e}^*, \theta_G^* = \argmin_{\theta_{c}, \theta_{a}, \theta_{E_{e}}, \theta_G} \mathcal{L}_2
\end{equation}
where $\theta_{c}$, $\theta_{a}$, and $\theta_{e}$ are the parameters of class, AU, and content encoders, respectively.

\subsubsection{Stage 2b -- Mapping of aleatoric uncertainty latent space to normal distribution}

For a motion sequence with given class and content labels, its class and content embeddings remain fixed according to their labels. However, these embeddings do not account for the variations that may occur within sequences that share the same class and content embeddings.
This is addressed with an AU latent embedding, somewhat analogous to having a slack variable. Additionally, we aim to synthesize new, diverse motion sequences even when the class and content labels are fixed. Since the distribution of the AU latent space is unknown, it must be mapped to a latent space with a known distribution from which we can sample multiple AU latents. To achieve this, we map the aleatoric uncertainty latent space to another latent space that follows a Gaussian distribution, using a VAE \cite{Kingma_2014_ICLR}, as shown in Figure \ref{fig:stage2}(b). 

The encoder and decoder of the VAE are neural networks with parameters $\theta$ and $\phi$, respectively. The encoder outputs are the mean and variance of a conditional Gaussian distribution $q_\theta(z|a)$, from which we can sample to obtain $z \,{\sim}\, q_{\theta}(z|a)$. The decoder takes $z$ as input, and it produces the parameters of the distribution $p_{\phi}(a|z)$. The loss function of the VAE is the negative evidence lower bound (ELBO) given by
\begin{align}\label{vae}
    \mathcal{L}_{\mathrm{VAE}}(\theta,\phi) = \sum_{i=1}^n& - \mathbb{E}_{z\sim q_{\theta}(z|a_i)}[\log p_{\phi}(a_i \mid z)] \notag \\
    &+\mathbb{KL}( q_{\theta}(z|a_i) \parallel p(z))
\end{align}
The first term is the reconstruction loss or the expected negative log-likelihood of the $i^{th}$ datapoint. The expectation is taken with respect to the encoder's distribution over the representations. This term encourages the decoder to learn to reconstruct the vector $a_i$ given its latent representation $z$. The second term can be interpreted as a regularizer and is the Kullback-Leibler divergence between the encoder's distribution $q_{\theta}(z|a_i)$ and $p(z)$. Here $p(z)$ is defined as $\mathcal{N}(0,I)$, so that the latent space distribution learned by the encoder is close to a standard normal distribution, which facilitates easy sampling of AU latents during synthesis.
 
By training the VAE on a dataset of AU latents, the encoder $E$ and the decoder $D$ parameters are optimized to learn a mapping from the given AU latent space to another latent space that follows a Gaussian distribution. During testing, a sample is taken from a Gaussian distribution $\mathcal{N}(0, I)$ and passed through the VAE decoder $D$ to generate the AU latent.

\subsubsection{Motion stylization with diverse motion generation}

\begin{figure}
  \centering
  \includegraphics[width=0.9\columnwidth]{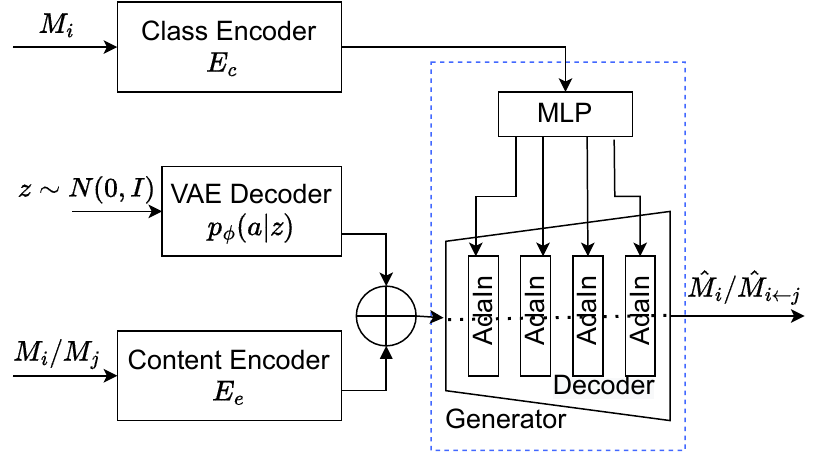}
   \caption{Motion stylization / Diverse motion sequence generation / Inference:  Diverse motion sequences from a given motion sequence can be generated by sampling multiple AU latents from Gaussian distribution. During inference, the motion sequence whose style is to be transferred is given as input to the class encoder, and the motion sequence whose motion is to be stylized is given as input to the content encoder.}
   \label{fig:inference}
\end{figure}

After training the D-LORD model, motion stylization is performed using two motion sequences, as shown in Figure \ref{fig:D-LORD}(b) and \ref{fig:inference}. The sequence whose style / identity is to be retained is provided to the Class Encoder $(E_{c})$, and the sequence representing the action to be performed is given to the Content Encoder $(E_e)$. A sample from a Gaussian distribution $\mathcal{N}(0, I)$ is passed through the VAE decoder $(D)$ to generate the AU latent, which captures aleatoric uncertainty and contributes to the diversity of the generated motions. The class latent (style) is then passed to an MLP, which maps the AdaIN parameters, adjusting the mean and variance of the decoder’s AdaIN layers for style adaptation. Finally, the AU latent is combined with the output of the Content Encoder $(E_e)$ and fed into the decoder $(G_{\theta_G})$ to generate the stylized motion sequence. This process is described as follows: 
\begin{equation}
  \hat{M}_{i \leftarrow j} = G_{\theta_G} \big( E_{c}(M_i), D_{\phi} (z \sim \mathcal{N}(0,I)), E_e(M_j) \big)
\end{equation}

For generating diverse motion sequences, the class and content latents are extracted by passing a motion sequence through the Class Encoder $(E_c)$ and Content Encoder $(E_e)$. Multiple AU latents are then sampled from a Gaussian distribution $\mathcal{N}(0, I)$ and decoded using the VAE decoder $D_{\phi}$. This process generates diverse motion sequences that preserve the specified class and content labels while incorporating variations in aleatoric uncertainty.

\section{Experiments and evaluations}\label{section:results}
This section first details the implementation, the network architecture of the D-LORD framework, and the datasets used in the experiments. We then report on various experiments conducted to evaluate the performance of the D-LORD framework. Qualitative and quantitative results on these datasets are discussed and compared with state-of-the-art algorithms.

\subsection{Implementation Details} \label{Implementation}
Our D-LORD motion style transfer framework is implemented using PyTorch, and the experiments were conducted on a PC equipped with an NVIDIA GeForce RTX 3090 GPU (24 GB) and Intel Core i9-10900K/3.70GHz CPU (64GB RAM).

\subsubsection{Network Architecture}\label{net_archi}
\begin{figure}
  \centering
  \includegraphics[width=0.9\columnwidth]{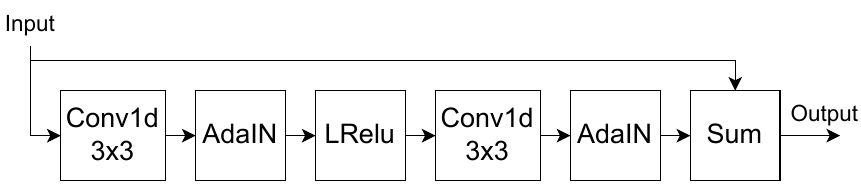}
   \caption{Residual Block}
   \label{fig:RB}
\end{figure}

The proposed framework uses the MLP and decoder networks in both Stage 1 and Stage 2, while the encoder networks are only used in Stage 2. The MLP network, comprising fully connected layers with Leaky ReLU activation, takes class embeddings as input to generate AdaIN parameters for the decoder's AdaIN layers. The decoder network features fully connected layers, residual blocks (see Figure \ref{fig:RB}), upsampling layers, and 1D convolution.
The output of the MLP network sets the parameters of the decoder's AdaIn layers. Then the decoder network concatenates the content embedding and AU embedding and processes them to reconstruct the input motion sequence. The residual blocks enhance motion quality while upsampling and 1D convolutions aid sequence generation. The encoder networks consist of six 1D convolution layers followed by three fully connected layers with Leaky ReLU activation, extracting local features and transforming them into embeddings. A VAE network maps the aleatoric uncertainty latents to a Gaussian distribution, enabling sampling for diverse motion generation.

Please refer to the Appendix for detailed information on the network architecture and hyperparameter tuning. An open-source implementation of the framework in PyTorch will be released soon.

\begin{figure*}[t]
\begin{center}
\begin{tabular}{cc}
\toprule
Class by Class & Content by Content \\ \hline
\includegraphics[scale=.3,valign=c]{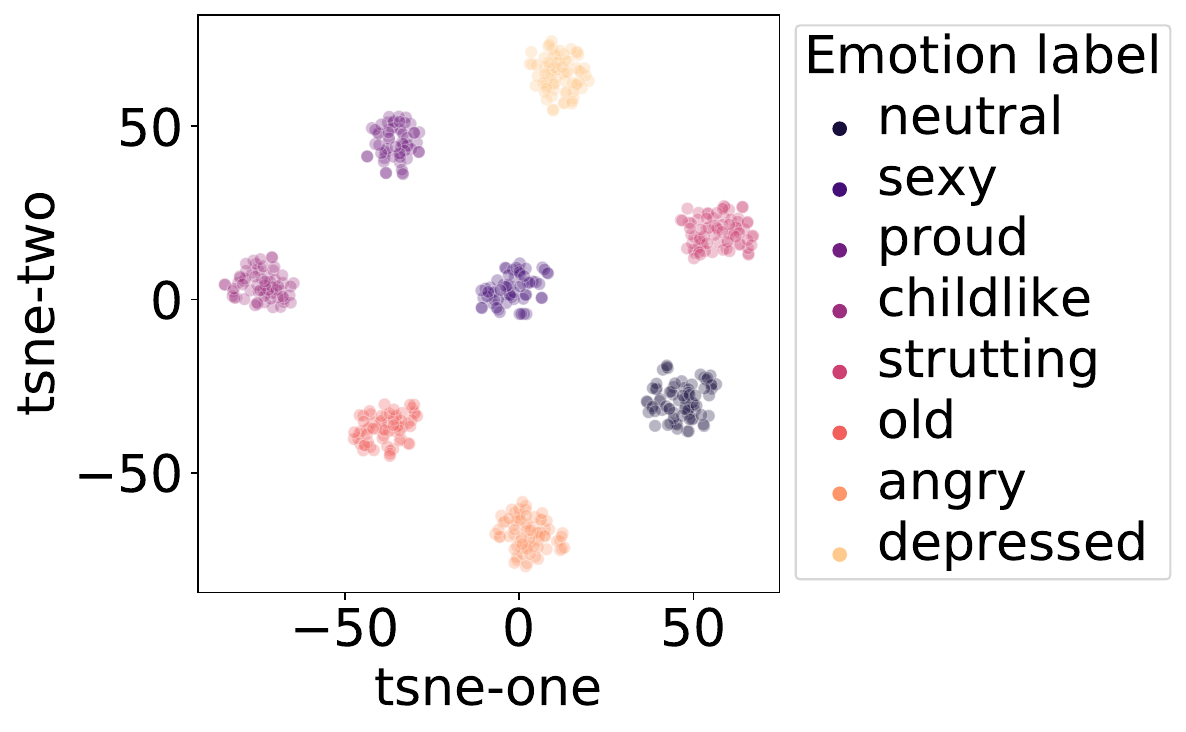} &  \includegraphics[scale=.3,valign=c]{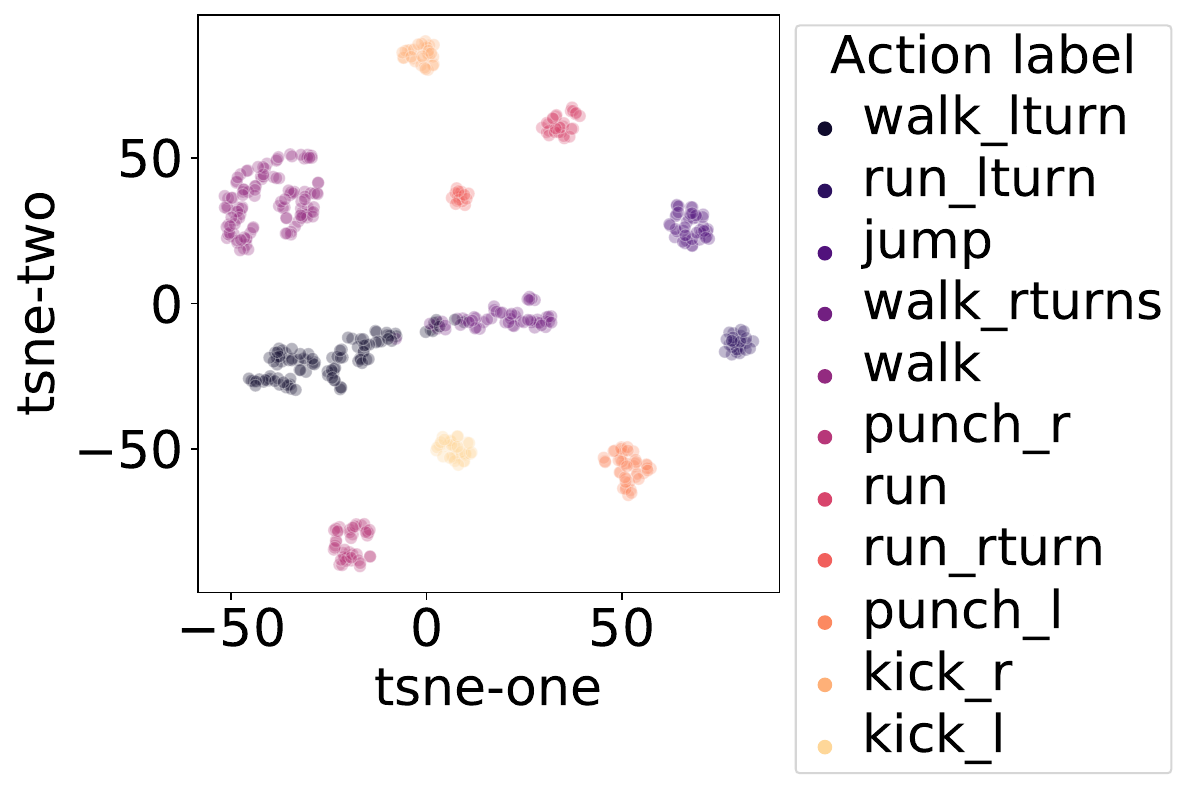}  \\ \hline
 \multicolumn{2}{c}{Xia dataset} \\ \hline
\includegraphics[scale=.3,valign=c]{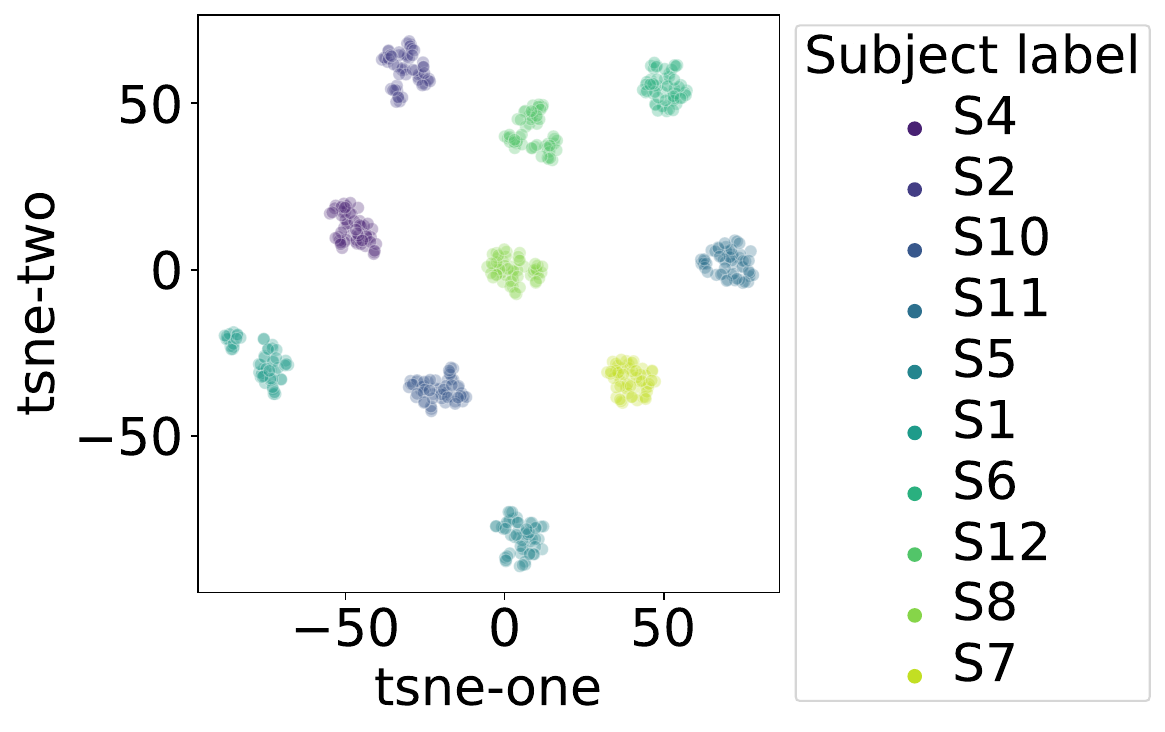} & \includegraphics[scale=.3,valign=c]{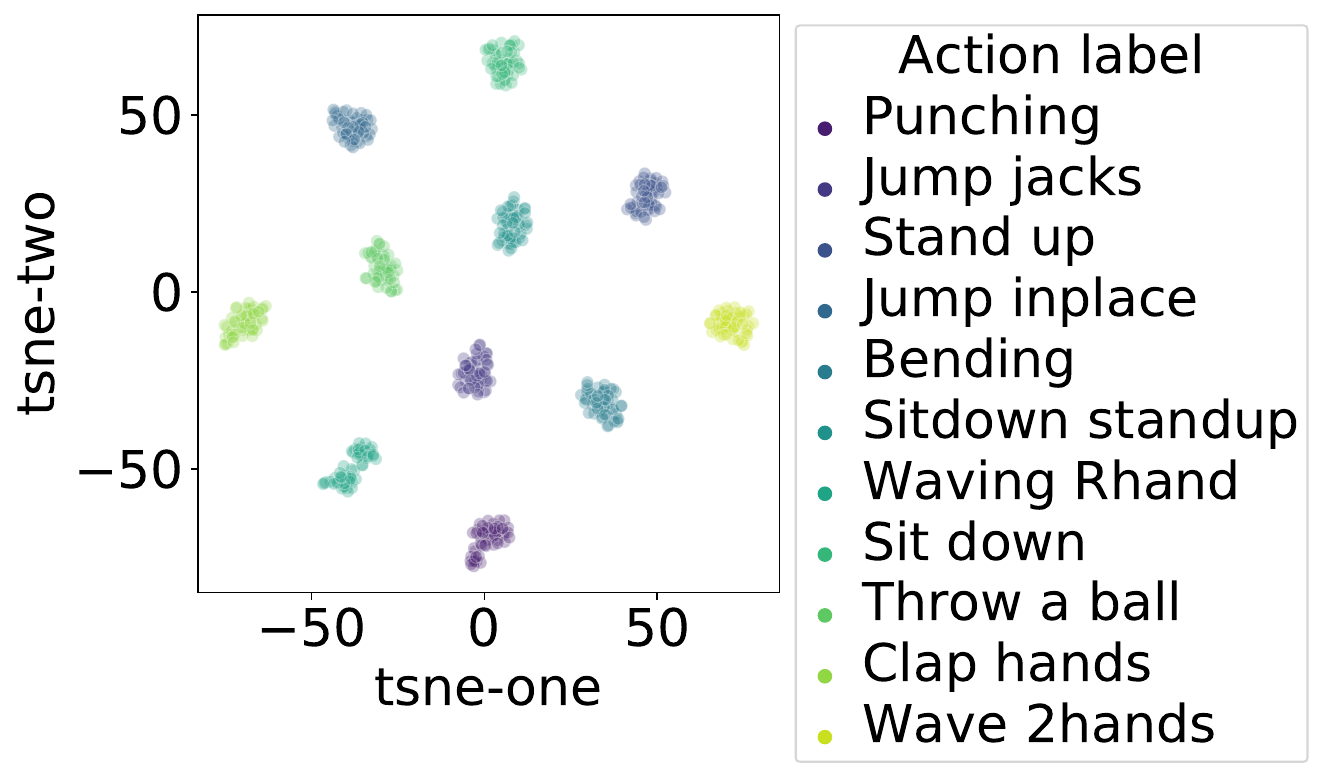} \\ \hline
 \multicolumn{2}{c}{MHAD dataset} \\ 
 \bottomrule
\end{tabular}
\end{center}
\caption{Class and content codes (128D) are projected onto a 2D space using t-SNE and plotted with respect to their respective labels. It can be seen that our framework learns to cluster the latent codes as a function of their label and does not require any additional triplet loss for clustering.}
\label{fig:condition1}
\end{figure*}

\subsubsection{Datasets}\label{dataset}
In this paper, we conducted experiments on three datasets: CMU Xia, Berkeley MHAD, and RRIS. For the Berkeley MHAD \cite{mhad} and RRIS \cite{liang2020asian} datasets, the class label is the subject ID, while for the CMU Xia dataset \cite{xia2015realtime}, it is the emotion label (e.g., angry). In all cases, the action label serves as the content label.

\emph{CMU Xia dataset} \cite{xia2015realtime}: It is publicly available and contains 572 motion sequences of 28 actions performed in 8 different emotions by a single actor. Each sequence has 3D positions of 21 joints and is resampled to 256 frames for standardization. Transitional actions are excluded, and some actions are grouped under common labels. The dataset is split into training (80$\%$) and test (20$\%$) sets.

\emph{Berkeley Multimodal Human Action Database} \cite{mhad}: It consists of 659 motion sequences from 12 subjects, with each subject performing 11 actions, repeated 5 times. The sequences capture 3D positions of 33 joints and are resampled to 256 frames for standardization. The training set includes all 11 actions and repetitions from 10 subjects, while the test set contains data from the remaining 2 subjects.

\emph{RRIS Ability Data} \cite{liang2020asian}: This dataset, obtained from the authors of \cite{liang2020asian}, contains 3,998 motion sequences performed by 200 individuals. It includes four lower limb tasks, automatically segmented, with each sequence resampled to 256 frames and 3D positions of 64 markers. The dataset is split into training, test, and validation sets: the training set includes sequences from 150 individuals, the test set contains sequences from 30 individuals, and the validation set includes sequences from 20 individuals.

\subsection{Latent Space Disentanglement Evaluation}

\begin{figure*}[t]
\begin{center}
\begin{tabular}{cc}
\toprule
 Class by Content & Content by Class \\ \hline
 \includegraphics[scale=.3,valign=c]{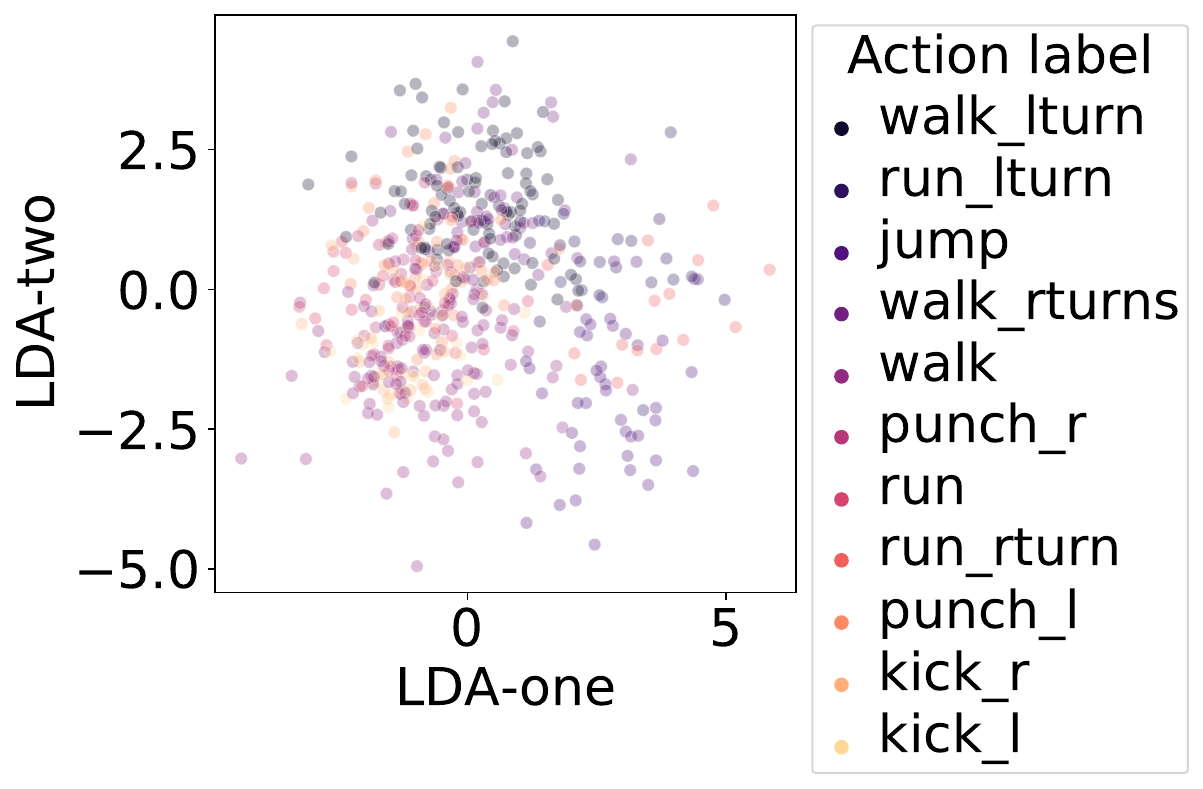}&  \includegraphics[scale=.3,valign=c]{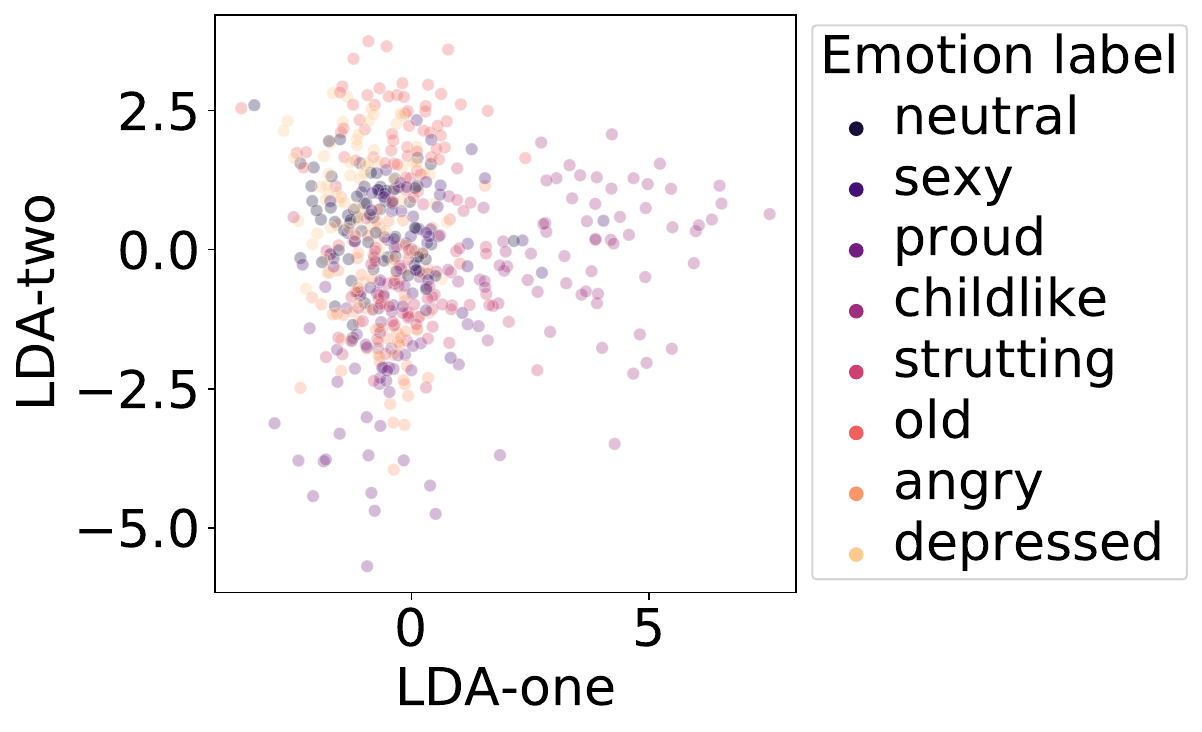} \\ \hline
 \multicolumn{2}{c}{Xia dataset} \\ \hline
 \includegraphics[scale=.3,valign=c]{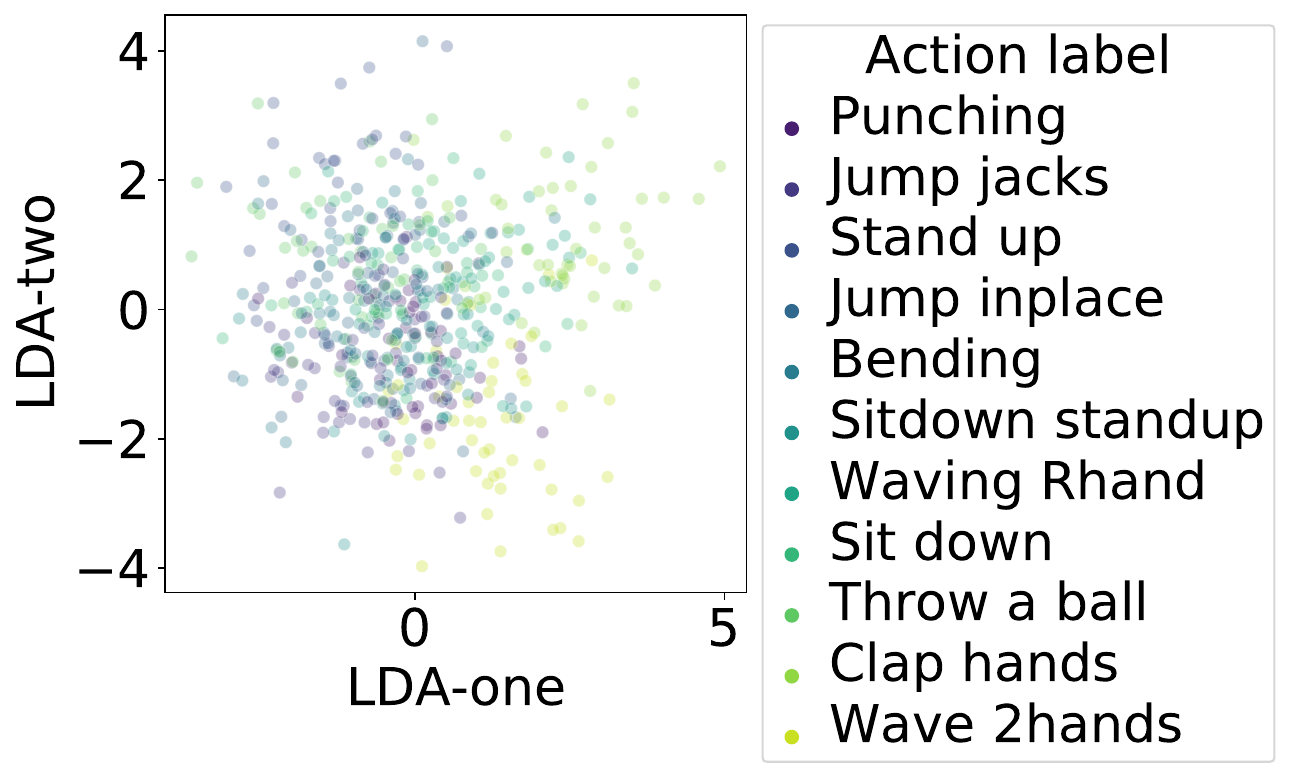} &  \includegraphics[scale=.3,valign=c]{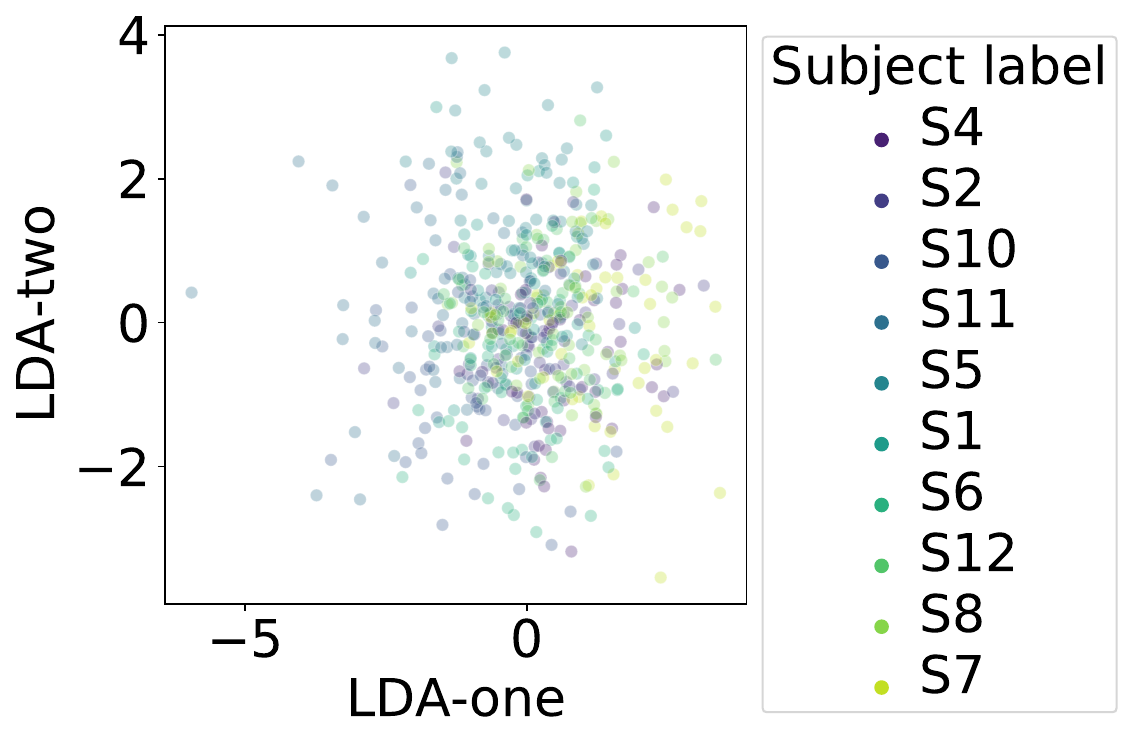}  \\ \hline
\multicolumn{2}{c}{MHAD dataset} \\ 
 \bottomrule
\end{tabular}
\end{center}
\caption{Column 1: An LDA classifier is trained for class codes and content labels. The trained classifier is used to project class codes (128D) in 2D  space and is plotted with respect to the content labels. Column 2: An LDA classifier is trained for content codes and class labels. The trained classifier was used to project content codes (128D) in 2D  space and plotted with respect to the class labels.}
\label{fig:condition2}
\end{figure*}

\begin{figure*}
        \centering
        \setlength{\extrarowheight}{0.2em}
        \begin{tabular}{|ccc|}
           \hline
            Input to Class encoder & Input to Content encoder & Output \\
             \hline 
              & Xia data-set & \\ 
              \includegraphics[scale=.45]{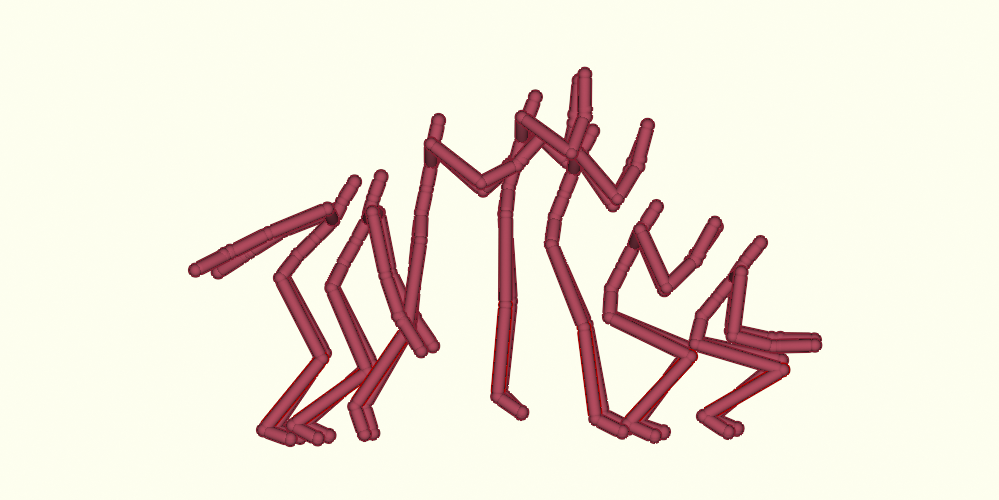} & \includegraphics[scale=.45]{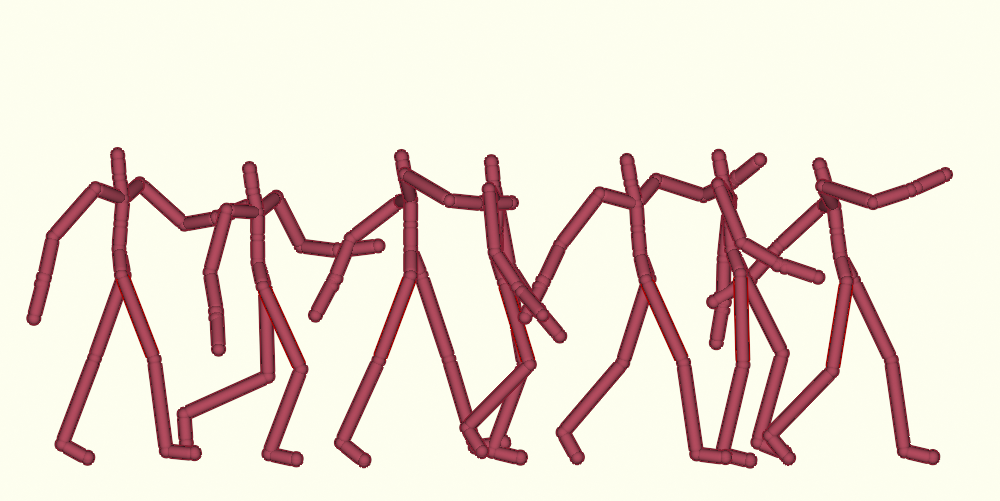} &\includegraphics[scale=.45]{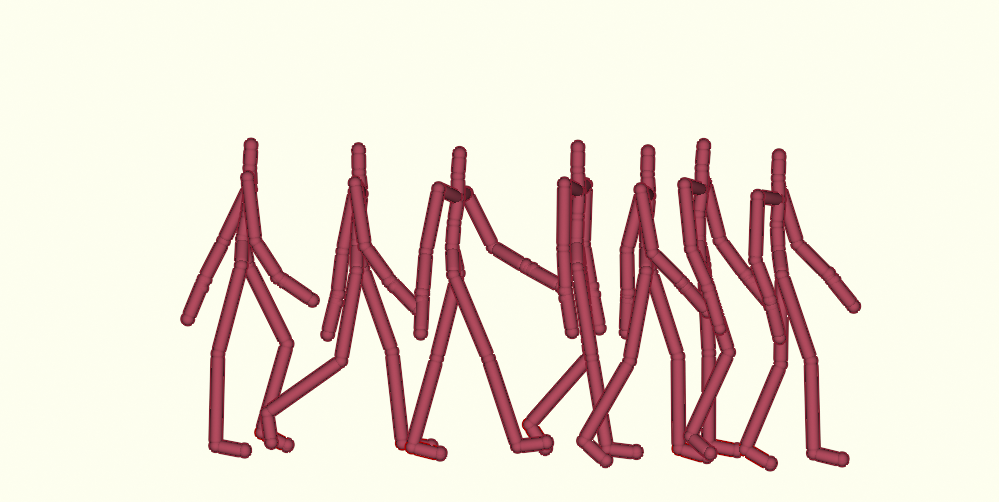}  \\
                 \textbf{Neutral} Jump & Proud \textbf{walk} &  \textbf{Neutral Walk}   \\ \hline 
                \bottomrule
             & MHAD data-set & \\ 
            \includegraphics[scale=.45]{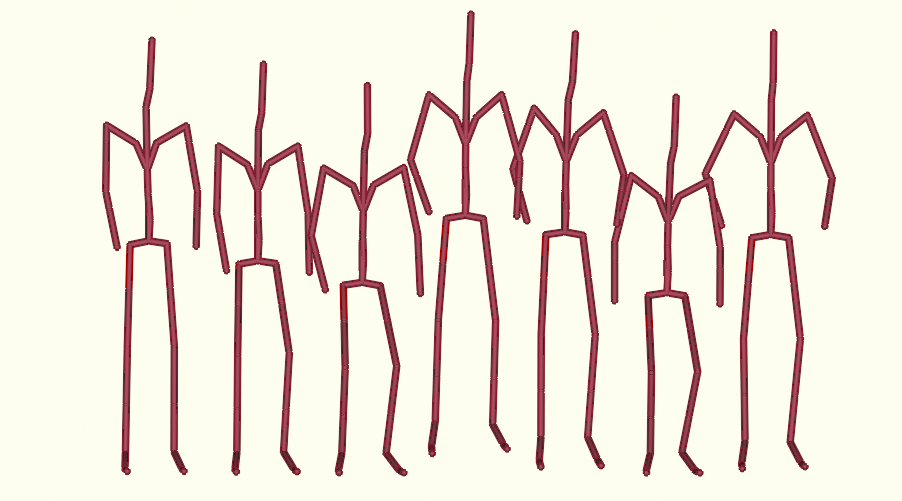} &\includegraphics[scale=.45]{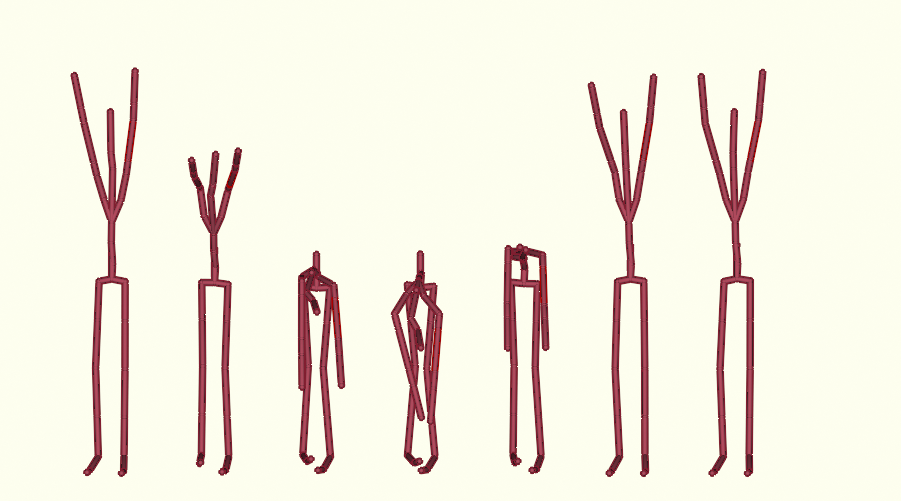}& \includegraphics[scale=.45]{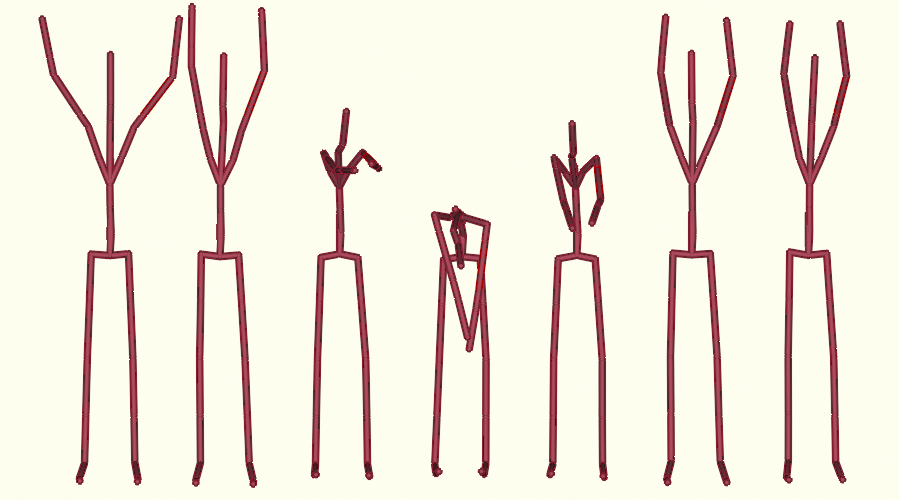}   \\ 
            \textbf{Identity $\boldsymbol{S_3}$}, Jump in place task  &  Identity ${S_{9}}$, \textbf{Bending task} & \textbf{Identity $\boldsymbol{S_3}$, Bending task}   \\ 
              \bottomrule
             & RRIS data-set & \\ 
            \includegraphics[scale=.5]{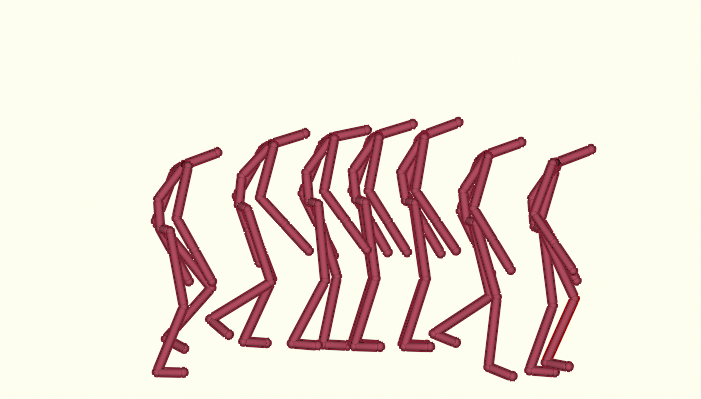} & \includegraphics[scale=.5]{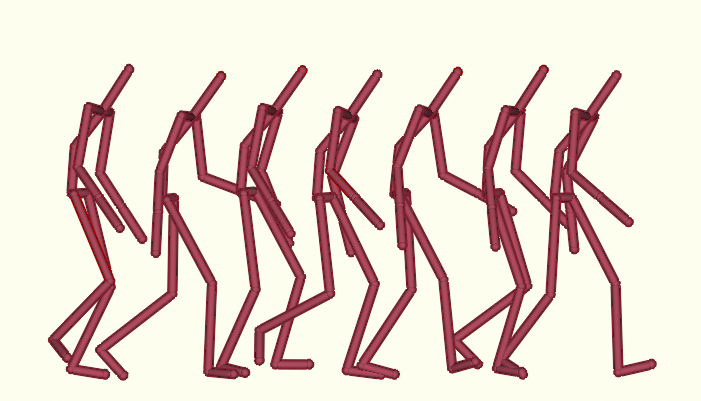} &\includegraphics[scale=.5]{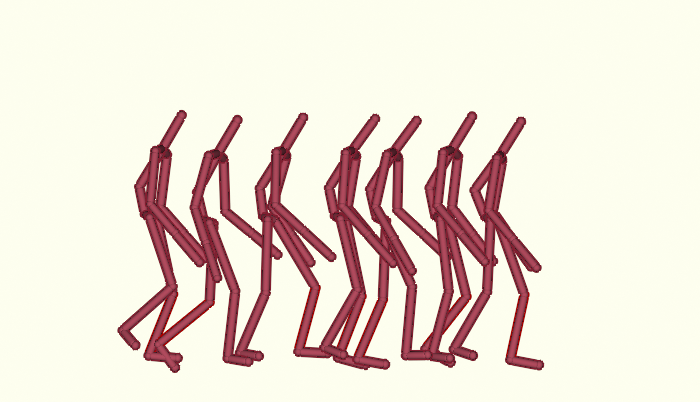}  \\
            \textbf{Identity $\boldsymbol{S_{11}}$}, Step-up task  &  Identity ${S_{101}}$, \textbf{Walk task}  &   \textbf{Identity $\boldsymbol{S_{11}}$, Walk task}  \\
            \bottomrule
        \end{tabular}
        \caption{D-LORD motion stylization results. The first row shows the results of motion style transfer on CMU Xia dataset. The second, and the third rows show the results of motion retargeting on MHAD, and RRIS datasets, respectively.}
        \label{fig:sample run}
    \end{figure*}

\setlength{\tabcolsep}{3pt}
\begin{table}
    \centering
    \begin{tabular}{ccccccccc}
    
      & EML1 & EML2 & EML3 & EML4  & EML5  & EML6 & EML7 & EML8  \\ 
      CT1 &  \textbf{0.032} & 0.543 & 0.542 & 0.488 & 0.554 & 0.537 & 0.542 & 0.521 \\ 
      CT2 &   0.544 & \textbf{0.029} & 0.533 & 0.547 & 0.577 & 0.517 & 0.523 & 0.54  \\ 
      CT3 & 0.543 & 0.533 & \textbf{0.027} & 0.513 & 0.57  & 0.527 & 0.555 & 0.572 \\ 
      CT4 & 0.488 & 0.547 & 0.513 & \textbf{0.031} & 0.548 & 0.497 & 0.471 & 0.482 \\ 
      CT5 & 0.555 & 0.577 & 0.57  & 0.549 & \textbf{0.027} & 0.57  & 0.567 & 0.577 \\ 
      CT6 & 0.537 & 0.517 & 0.526 & 0.496 & 0.57  & \textbf{0.033} & 0.498 & 0.531 \\ 
      CT7 & 0.542 & 0.523 & 0.554 & 0.471 & 0.567 & 0.499 & \textbf{0.032} & 0.566 \\ 
      CT8 & 0.521 & 0.54  & 0.572 & 0.482 & 0.577 & 0.531 & 0.566 & \textbf{0.03}  \\ 
    \end{tabular}
    \caption{Average distance between Class latents centroids and the latent codes that share the same class label for XIA-dataset. EML: Emotion label, CT: Centroid }
    \label{tab:xia_class}
\end{table}

\setlength{\tabcolsep}{2.5pt}
\begin{table}
    \centering
    \begin{tabular}{ccccccccccccc}
         & AL1 & AL2 & AL3 & AL4  & AL5  & AL6 & AL7 & AL8 & AL9 & AL10 & AL11 \\ 
 CT1 &  \textbf{0.02} & 0.54 & 0.55 & 0.50 & 0.61 & 0.62 & 0.59 & 0.74 & 0.70 & 0.78 & 0.78 \\  
        CT2 & 0.54 & \textbf{0.07} & 0.32 & 0.50 & 0.47 & 0.49 & 0.48 & 0.60 & 0.60 & 0.61 & 0.60 \\ 
        CT3 & 0.55 & 0.32 & \textbf{0.08} & 0.52 & 0.48 & 0.49 & 0.46 & 0.60 & 0.61 & 0.63 & 0.59 \\ 
        CT4 & 0.50 & 0.50 & 0.52 & \textbf{0.05} & 0.60 & 0.55 & 0.59 & 0.70 & 0.71 & 0.74 & 0.73 \\ 
        CT5 & 0.61 & 0.47 & 0.49 & 0.59 & \textbf{0.04} & 0.54 & 0.54 & 0.63 & 0.66 & 0.74  & 0.65 \\ 
        CT6 & 0.62 & 0.49 & 0.50 & 0.55 & 0.53 & \textbf{0.05} & 0.55 & 0.65 & 0.64 & 0.66  & 0.63 \\ 
        CT7 & 0.60 & 0.48 & 0.47 & 0.59 & 0.54 & 0.55 & \textbf{0.04} & 0.63  & 0.67 & 0.73 & 0.64 \\ 
        CT8 & 0.74 & 0.60 & 0.61 & 0.71 & 0.63 & 0.65 & 0.63 & \textbf{0.02} & 0.66 & 0.72 & 0.72 \\ 
        CT9 & 0.71 & 0.61 & 0.62 & 0.71 & 0.66 & 0.64 & 0.68 & 0.66 & \textbf{0.02} & 0.74 & 0.73 \\ 
        CT10 & 0.80 & 0.62 & 0.64 & 0.74 & 0.74 & 0.66 & 0.73 & 0.74 & 0.74 & \textbf{0.03} & 0.73 \\ 
        CT11 & 0.78 & 0.60 & 0.60 & 0.73  & 0.65 & 0.63  & 0.64 & 0.71 & 0.73 & 0.73 & \textbf{0.04} \\  
     \end{tabular}
    \caption{Average distance between Content latents centroids and the latent codes that share the same content label for XIA-dataset. AL: Action label, CT: Centroid }
    \label{tab:xia_content}
\end{table}

\setlength{\tabcolsep}{6pt}
\begin{table*}
    \centering
    \begin{tabular}{cccccccccccc}
    & SL1 & SL2 & SL3 & SL4 & SL5 & SL6 & SL7 & SL8 & SL9 & SL10 \\ 
    CT1 &    \textbf{0.0024} & 0.3131 & 0.3636 & 0.3200 & 0.3114 & 0.2784 & 0.2754 & 0.2835 & 0.2947 & 0.2212 \\ 
    CT2 &    0.3131 & \textbf{0.0022} & 0.3574 & 0.2759 & 0.2532 & 0.3557 & 0.3000 & 0.2595 & 0.2404 & 0.2341 \\ 
    CT3 &    0.3636 & 0.3574 & \textbf{0.0018} & 0.3728 & 0.3242 & 0.3298 & 0.3443 & 0.3629 & 0.3513 & 0.3380 \\ 
    CT4 &    0.3200 & 0.2759 & 0.3728 & \textbf{0.0020} & 0.2634 & 0.3711 & 0.2699 & 0.2671 & 0.2360 & 0.2610 \\ 
    CT5 &    0.3114 & 0.2532 & 0.3242 & 0.2634 & \textbf{0.0022} & 0.3205 & 0.2587 & 0.2621 & 0.2358 & 0.2504 \\ 
    CT6 &    0.2784 & 0.3557 & 0.3298 & 0.3711 & 0.3205 & \textbf{0.0018} & 0.2741 & 0.3490 & 0.3408 & 0.2808 \\ 
    CT7 &    0.2754 & 0.3000 & 0.3443 & 0.2699 & 0.2587 & 0.2741 & \textbf{0.0021} & 0.2645 & 0.2657 & 0.2490 \\ 
    CT8 &    0.2835 & 0.2595 & 0.3629 & 0.2671 & 0.2621 & 0.3490 & 0.2645 & \textbf{0.0020} & 0.2221 & 0.2348 \\ 
    CT9 &    0.2947 & 0.2404 & 0.3513 & 0.2360 & 0.2358 & 0.3408 & 0.2657 & 0.2221 & \textbf{0.0023} & 0.2373 \\ 
    CT10 &    0.2212 & 0.2341 & 0.3380 & 0.2610 & 0.2504 & 0.2808 & 0.2490 & 0.2348 & 0.2373 & \textbf{0.0024} \\ 
\end{tabular}
    \caption{Average distance between Class latents centroids and the latent codes that share the same class label for MHAD-dataset. SL: Subject label, CT: Centroid }
    \label{tab:mhad_class}
\end{table*}

\setlength{\tabcolsep}{4pt}
\begin{table*}
    \centering
    \begin{tabular}{ccccccccccccc}
    &  AL1 & AL2 & AL3 & AL4 & AL5 & AL6 & AL7 & AL8 & AL9 & AL10 & AL11 \\
    CT1 &   \textbf{2.000e-04} & 4.184e-01 & 4.041e-01 & 4.041e-01 & 4.890e-01 & 4.622e-01 & 4.720e-01 & 4.318e-01 & 4.538e-01 & 4.354e-01 & 4.200e-01 \\ 
    CT2 &    4.184e-01 & \textbf{2.000e-04} & 4.212e-01 & 4.398e-01 & 4.484e-01 & 4.847e-01 & 5.028e-01 & 4.406e-01 & 4.963e-01 & 4.748e-01 & 4.402e-01 \\ 
    CT3 & 4.041e-01 & 4.212e-01 & \textbf{2.000e-04} & 4.170e-01 & 4.872e-01 & 4.995e-01 & 4.680e-01 & 4.336e-01 & 4.448e-01 & 4.795e-01 & 4.473e-01 \\ 
    CT4 &    4.041e-01 & 4.398e-01 & 4.170e-01 & \textbf{3.000e-04} & 5.072e-01 & 4.884e-01 & 4.329e-01 & 4.321e-01 & 4.770e-01 & 4.880e-01 & 4.400e-01 \\ 
    CT5 &    4.890e-01 & 4.484e-01 & 4.872e-01 & 5.072e-01 & \textbf{2.000e-04} & 5.109e-01 & 5.523e-01 & 4.639e-01 & 5.279e-01 & 5.251e-01 & 5.204e-01 \\ 
    CT6 &    4.622e-01 & 4.847e-01 & 4.995e-01 & 4.884e-01 & 5.109e-01 & \textbf{3.000e-04} & 5.009e-01 & 5.047e-01 & 4.975e-01 & 4.898e-01 & 4.976e-01 \\ 
    CT7 &    4.720e-01 & 5.028e-01 & 4.680e-01 & 4.329e-01 & 5.523e-01 & 5.009e-01 & \textbf{3.000e-04} & 4.638e-01 & 4.798e-01 & 4.822e-01 & 4.748e-01 \\ 
    CT8 &    4.318e-01 & 4.406e-01 & 4.336e-01 & 4.321e-01 & 4.639e-01 & 5.047e-01 & 4.638e-01 & \textbf{3.000e-04} & 4.415e-01 & 4.526e-01 & 4.094e-01 \\ 
    CT9 &    4.538e-01 & 4.963e-01 & 4.448e-01 & 4.770e-01 & 5.279e-01 & 4.975e-01 & 4.798e-01 & 4.415e-01 & \textbf{2.000e-04} & 4.463e-01 & 4.683e-01 \\ 
    CT10 &    4.354e-01 & 4.748e-01 & 4.795e-01 & 4.880e-01 & 5.251e-01 & 4.898e-01 & 4.822e-01 & 4.526e-01 & 4.463e-01 & \textbf{2.000e-04} & 4.625e-01 \\ 
    CT11 &    4.200e-01 & 4.402e-01 & 4.473e-01 & 4.400e-01 & 5.204e-01 & 4.976e-01 & 4.748e-01 & 4.094e-01 & 4.683e-01 & 4.625e-01 & \textbf{1.000e-04} \\ 
    \end{tabular}
    \caption{Average distance between Content latents centroids and the latent codes that share the same content label for MHAD-dataset. AL: Action label, CT: Centroid }
    \label{tab:mhad_content}
\end{table*}

We evaluated D-LORD's ability to disentangle class and content information based on two conditions: \begin{itemize}
\item Class and content codes should only contain information related to their respective labels. \item No content information should leak into the class code and vice versa. 
\end{itemize} 
To verify the first condition, we projected the class and content codes into a 2D space using t-SNE (t-distributed stochastic neighbor embedding) \cite{Vandermaaten_2008a_tSNE_JMLR} and plotted them according to their labels \ref{fig:condition1}. The t-SNE visualization showed well-clustered class and content codes, indicating successful disentanglement of the labels by the network.

To further validate this, we calculated the centroids of the class codes for each label and computed the average distance between these centroids and the corresponding latent codes. As shown in Table \ref{tab:xia_class} and Table \ref{tab:mhad_class}, the average distance between the latents and their respective class centroids was significantly smaller than the distance to other class centroids. This confirms that the network effectively captured class-specific information. The same approach was applied to the content latents, with results in Table \ref{tab:xia_content} and Table \ref{tab:mhad_content}, further confirming the network's ability to distinguish between content labels.

To test the second condition, we evaluated the potential leakage of content information into the class codes. We trained a Linear Discriminant Analysis (LDA) classifier using the class codes as inputs and the content labels as outputs. If content information were present in the class codes, the LDA projection in a 2D space, plotted with respect to content labels (figure \ref{fig:condition2}), would show distinct clustering. However, no such clustering was observed, confirming that the class codes do not contain content information. Similarly, we applied the same test to the content codes, and the absence of clustering based on class labels further verified that content codes remain independent of class information

Overall, the evaluation showed that D-LORD effectively disentangles class and content information from motion sequences. Additional results on the RRIS dataset, provided in the supplementary material, further confirm these findings.

\subsection{Visual results} 
 
 \begin{figure*}
\centering
\setlength{\extrarowheight}{0.2em}
\begin{tabular}{|cccc|}
\hline
 & \textbf{Neutral} Kick & & Proud \textbf{Jump} \\
Input to the Class encoder  & \includegraphics[scale=0.41,valign=c]{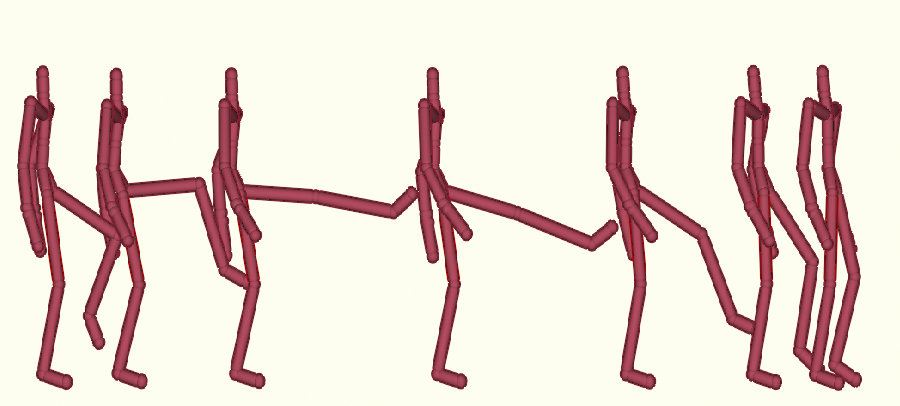} & Input to the Content encoder & \includegraphics[scale=0.41,valign=c]{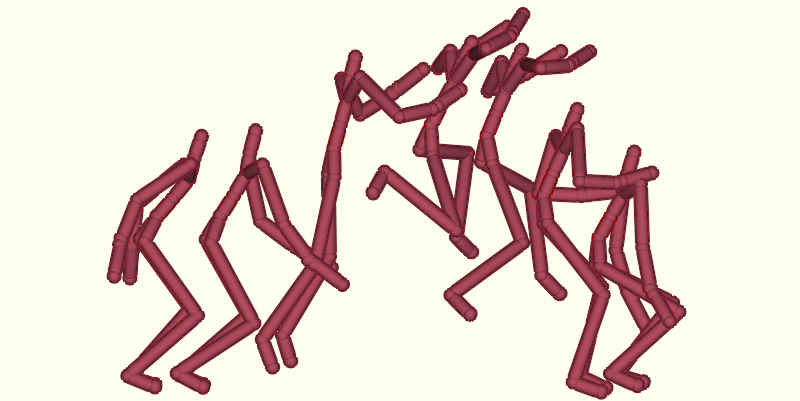}   \\
\hline 
   Output - \textbf{Neutral Jump} & & & \\ 
              Aberman et al.\ \cite{aberman2020} & Park et al.\ \cite{park2021diverse} & D-LORD & Ground Truth \\
               \includegraphics[scale=.41]{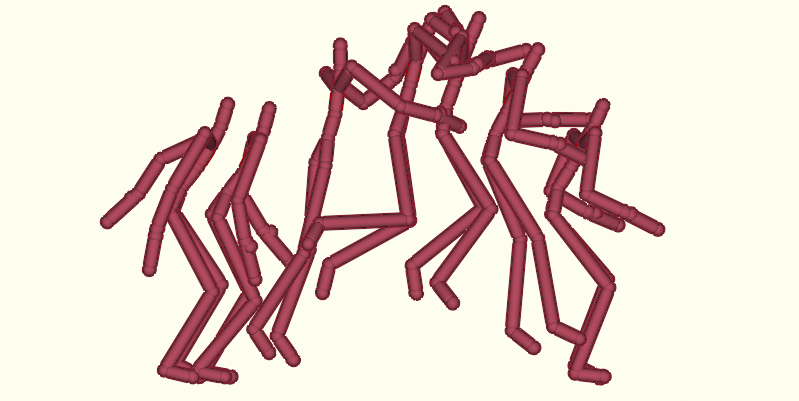} & \includegraphics[scale=.41]{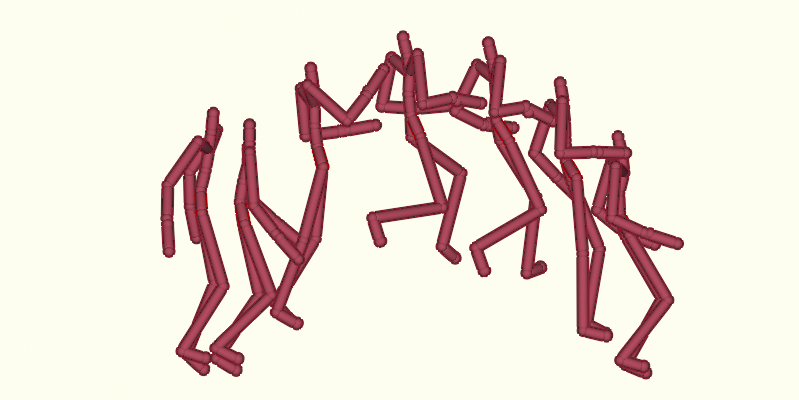} &\includegraphics[scale=.41]{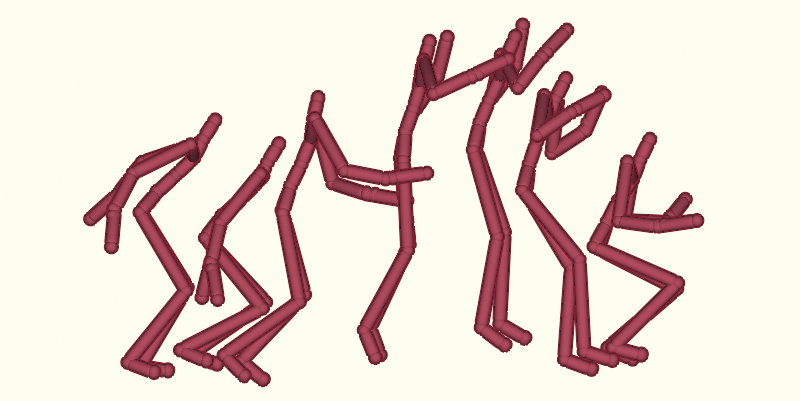} & \includegraphics[scale=.41]{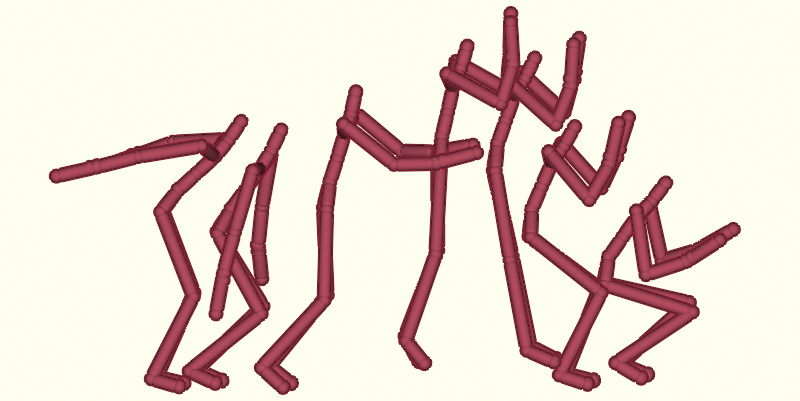}  \\ \hline 
\bottomrule
 &\textbf{Proud} Jump  & &  Neutral \textbf{Kick} \\
Input to the Class encoder   & \includegraphics[scale=0.41,valign=c]{fig/proud_jump.png} & Input to the Content encoder & \includegraphics[scale=0.41,valign=c]{fig/neutral_kick.png}   \\
\hline 
   Output - \textbf{Proud Kick} & & & \\ 
              Aberman et al.\ \cite{aberman2020} & Park et al.\ \cite{park2021diverse} & D-LORD & Ground Truth \\
               \includegraphics[scale=.41]{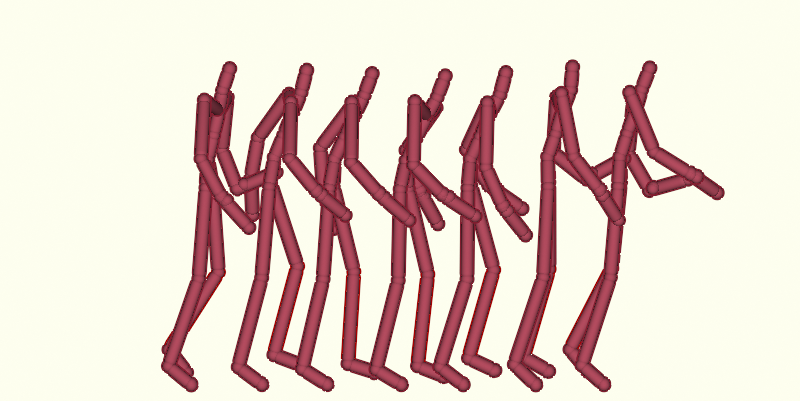} &\includegraphics[scale=.41]{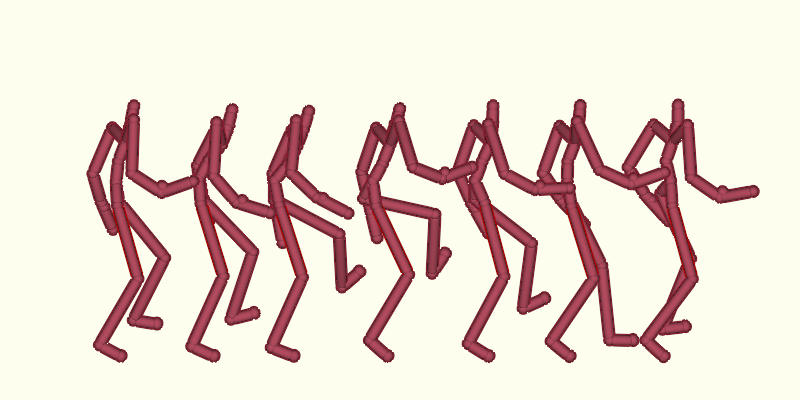}& \includegraphics[scale=.41]{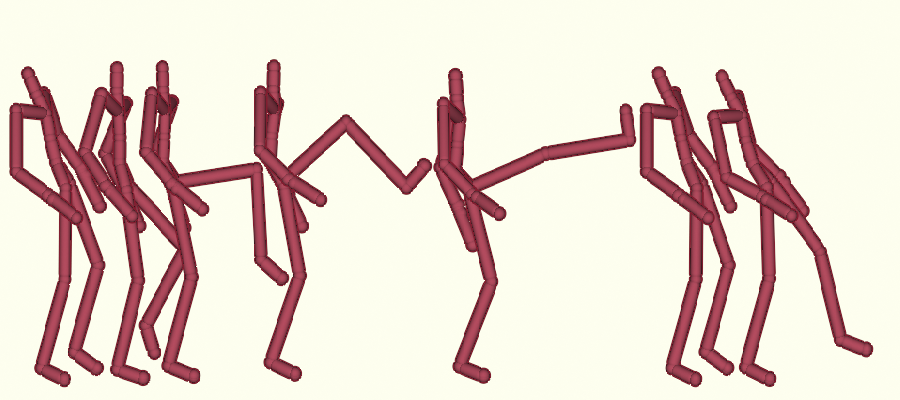} & \includegraphics[scale=.41]{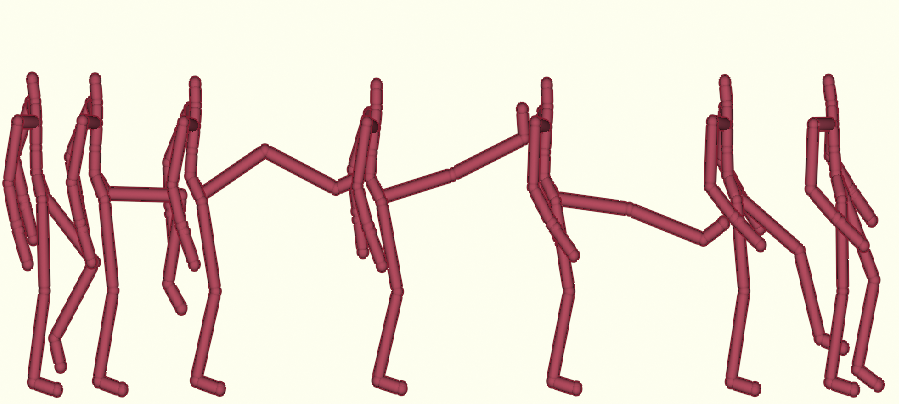} \\ \hline 
\end{tabular}
\caption{Qualitative comparison of our D-LORD to  Aberman et al.\ \cite{aberman2020} and Park et al.\ \cite{park2021diverse} on the CMU Xia dataset \cite{xia2015realtime}. Results were generated for two different inputs, as shown in rows 1 and 3. Corresponding generated output by Aberman et al., Park et al., ours, and the ground truth are shown in rows 2 and 4. A fixed set of keyframes was chosen for each motion to demonstrate the results. }
\label{fig:com_ms}
    \end{figure*}

Figure \ref{fig:sample run} shows the results of motion stylization experiments conducted on three datasets. In all cases, the AU latent was sampled from a Gaussian distribution.

In the first experiment, emotion transfer (motion style transfer) was applied to the CMU Xia dataset. The class encoder received a \textit{Neutral Jump} motion sequence, while the content encoder received a \textit{Proud Walk} motion sequence. The class encoder extracted the neutral style, and the content encoder extracted the walking motion. The generator then produced a \textit{Neutral Walk} motion sequence, successfully changing the style from proud to neutral (see first row in Figure \ref{fig:sample run}). The proud walk, characterized by a bent spine and raised shoulders, was transformed into a neutral walk with a straight spine and unraised shoulders.

The second experiment focused on identity transfer (motion retargeting) using the MHAD dataset. The class encoder received a \textit{Jump-in-place} motion sequence performed by \textit{Identity} $S_3$, while the content encoder received a \textit{Bending} motion sequence performed by \textit{Identity} $S_9$. The class encoder extracted the identity $S_3$ while the content encoder extracted the bending motion. The generator then produced a \textit{Bending} sequence performed by \textit{Identity} $S_3$. As shown in the second row of Figure \ref{fig:sample run}, the generator successfully transferred the bending motion from $S_9$ to $S_3$, which is taller than $S_9$.

In the third experiment, identity transfer was applied to the RRIS dataset. The class encoder received a \textit{Step-up}  motion sequence from \textit{Identity} $S_{11}$, and the content encoder received a \textit{Walk}  motion sequence from \textit{Identity} $S_{101}$. The class encoder extracted the identity $S_{11}$ while the content encoder extracted the walking motion. The generator then produced a \textit{Walk} sequence performed by \textit{Identity} $S_{11}$. As shown in the third row of Figure \ref{fig:sample run}, the generator transferred both the identity and gait length of $S_{11}$ to the walking sequence, changing the identity from $S_{101}$ to $S_{11}$, who is shorter.

These results demonstrate the effectiveness of the D-LORD framework in manipulating the emotion and identity of motion sequences based on the class and content inputs.

\subsection{Qualitative and Quantitative Comparison} 
\subsubsection{Motion style transfer}
\begin{figure*}
\centering
\setlength{\extrarowheight}{0.2em}
\begin{tabular}{cc|cc}
\hline
 \multicolumn{2}{c}{Inputs} &  \multicolumn{2}{c}{Outputs} \\  \hline 
 Class encoder &   Content encoder & D-LORD & LORD \\ \hline
 \includegraphics[scale=0.45]{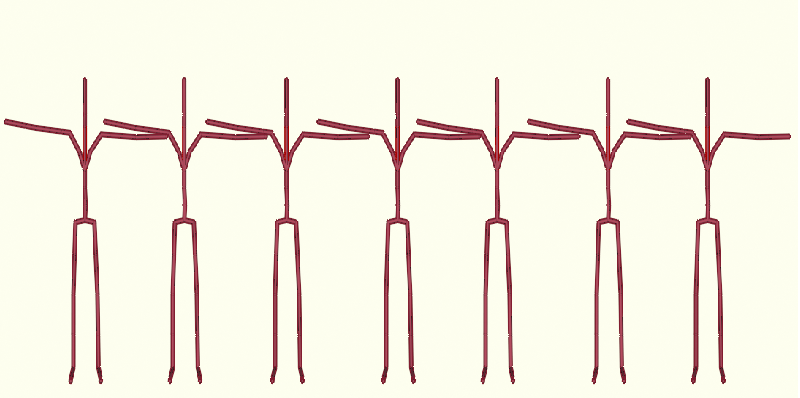} & \includegraphics[scale=0.45]{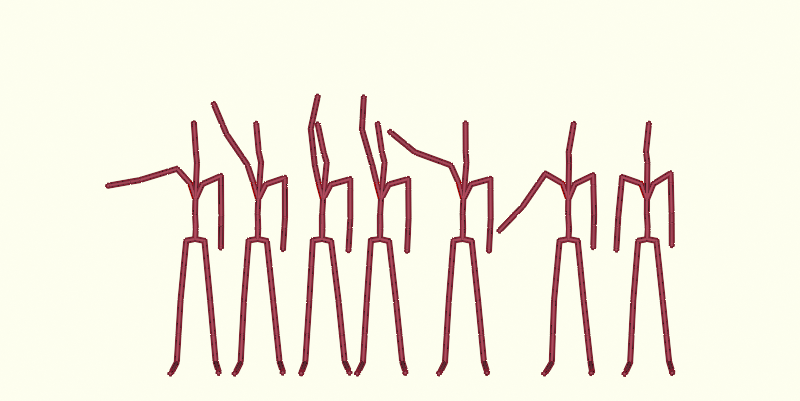} 
 & \includegraphics[scale=.45]{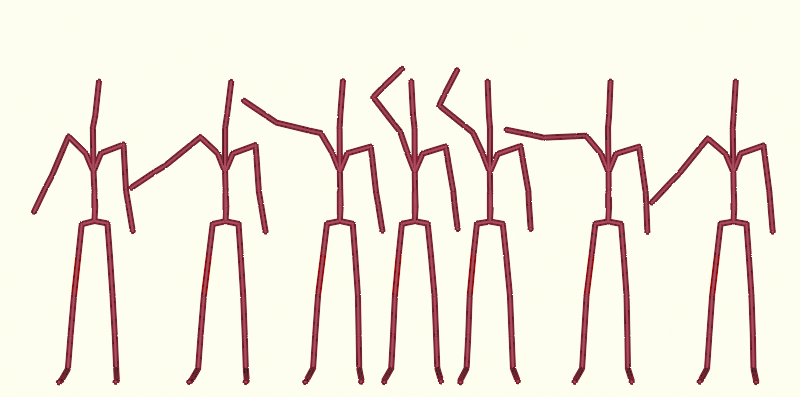} & \includegraphics[scale=.45]{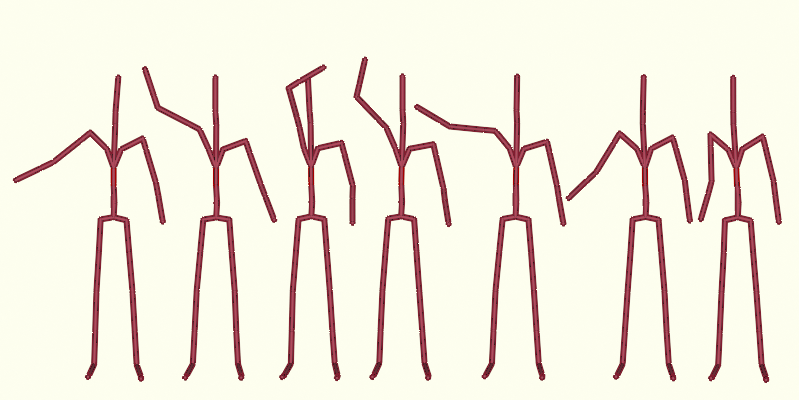}  \\
  Identity ${S_3}$, T-pose &  Identity ${S_9}$, Waving right hand  &  \multicolumn{2}{c}{Identity ${S_3}$,  Waving right hand }  \\ \hline 
 T-pose & Source motion   &  \multicolumn{2}{c}{Villegas et al.\ \cite{villegas2018neural}} \\ \hline
 \includegraphics[scale=0.45]{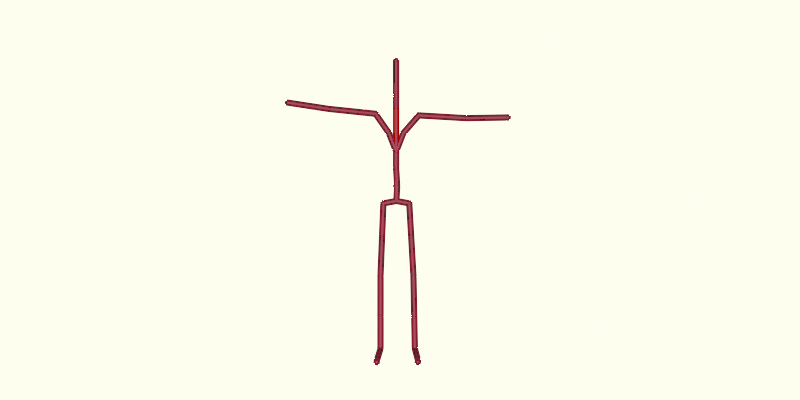} & \includegraphics[scale=0.45]{fig/mhad_in2.png} 
 &   \multicolumn{2}{c}{\includegraphics[scale=.45]{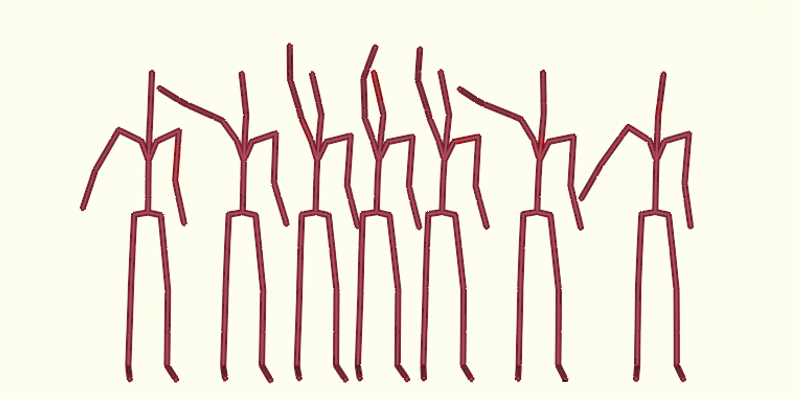}} \\
  Identity ${S_3}$, T-pose &  Identity ${S_9}$, Waving right hand  &  \multicolumn{2}{c}{Identity ${S_3}$,  Waving right hand }  \\ \hline \bottomrule
  Class encoder &   Content encoder & D-LORD & LORD \\ \hline
 \includegraphics[scale=0.45]{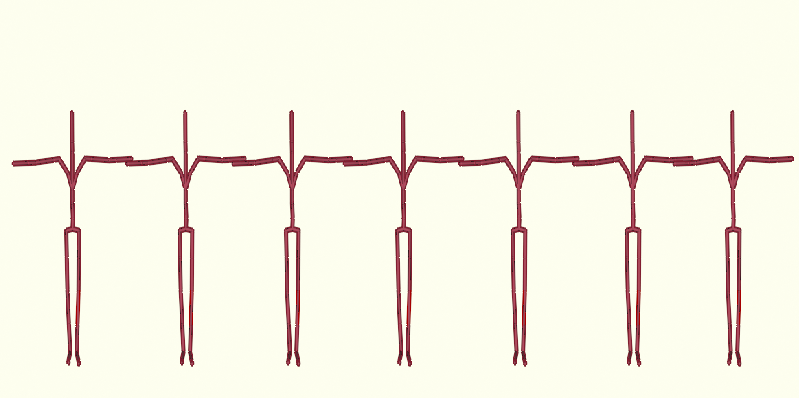} & \includegraphics[scale=0.45]{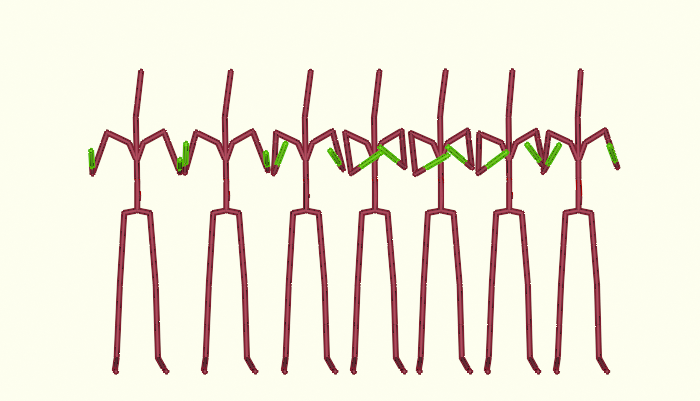} 
 & \includegraphics[scale=.45]{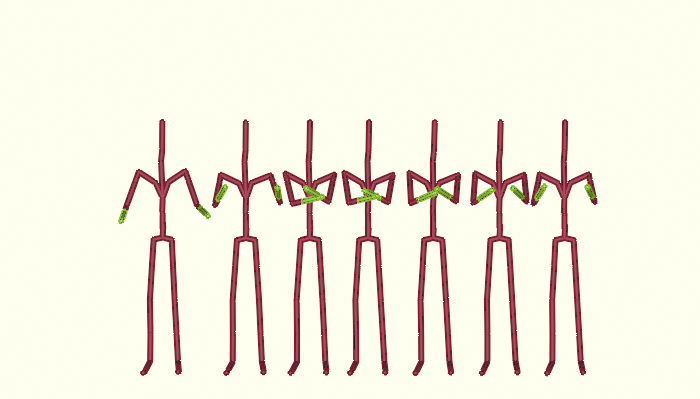} & \includegraphics[scale=.45]{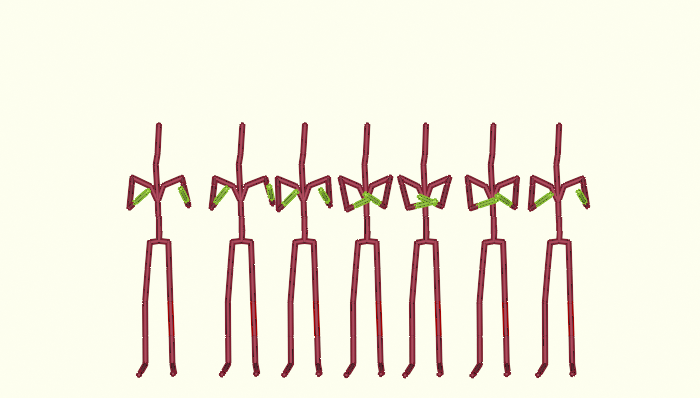}  \\ Identity ${S_9}$, T-pose &  Identity ${S_3}$, Clapping hands  &  \multicolumn{2}{c}{Identity ${S_9}$,   Clapping hands }  \\ \hline 
 T-pose & Source motion   &   \multicolumn{2}{c}{Villegas et al.\ \cite{villegas2018neural}} \\ \hline
 \includegraphics[scale=0.45]{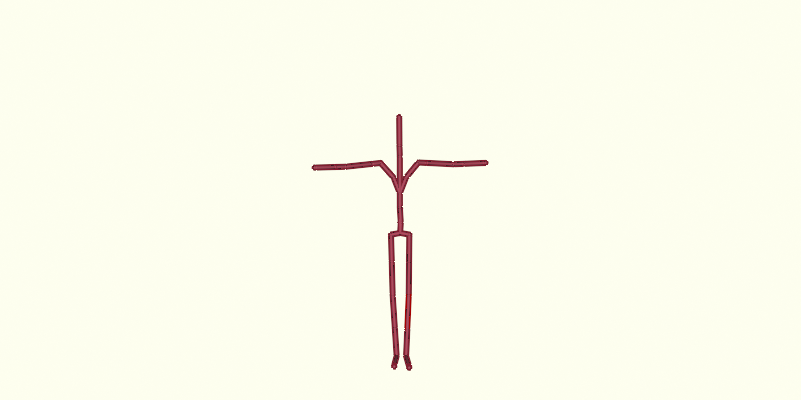} & \includegraphics[scale=0.45]{fig/mhad_in22.png} 
 &  \multicolumn{2}{c}{\includegraphics[scale=.45]{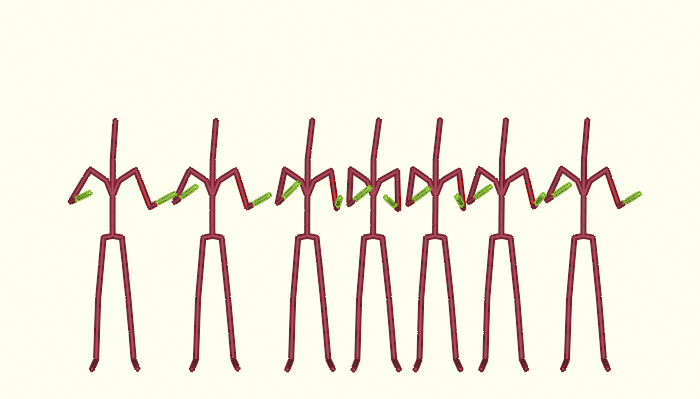}} \\
 Identity ${S_9}$, T-pose &  Identity ${S_3}$, Clapping hands  &  \multicolumn{2}{c}{Identity ${S_9}$,   Clapping hands }  \\ \hline 
\end{tabular}
\caption{Qualitative comparison of our methods to Villegas et al.\ \cite{villegas2018neural} on the MHAD dataset. We provide the results of our two methods (1) D-LORD, and (2) LORD with bone consistency loss.}
\label{fig:com_mr}
    \end{figure*}

We conducted a qualitative and quantitative comparison of our D-LORD framework with two state-of-the-art methods: Aberman et al.\ \cite{aberman2020} and Park et al.\ \cite{park2021diverse}, using the CMU Xia dataset. The evaluation focused on three main aspects: generation quality, style transfer, and content preservation. It is important to note that the methods proposed by Aberman et al.\ \cite{aberman2020} and Park et al.\ \cite{park2021diverse} are tailored for specific class labels, making them unsuitable for direct comparison on the other datasets used for motion retargeting in our experiments.

\emph{Qualitative evaluation -} Figure \ref{fig:com_ms} shows a visual comparison of D-LORD with existing methods. In the first experiment, using \textit{Neutral Kick} as the class input and \textit{Proud Jump} as the content input, the expected output is \textit{Neutral Jump}. As seen in row 2 of Figure \ref{fig:com_ms}, D-LORD generated a motion sequence closely matching the ground truth, while Park et al.\ \cite{park2021diverse} failed to transfer the neutral style effectively. In the second experiment, with swapped inputs, the output should be \textit{Proud Kick}. In row 4 of Figure \ref{fig:com_ms}, both existing methods struggled to preserve the source motion, failing to retain the kicking motion. However, D-LORD successfully maintained the source motion and accurately transferred the proud style. The existing methods underperformed, especially when the input motions differed significantly. In contrast, D-LORD effectively disentangles style and content, generating visually high-quality motion sequences.

\setlength{\tabcolsep}{8pt}
\begin{table}
\centering
\begin{tabular}{ ccccc} 
 \hline
 \multirow{2}{*}{Methods} & \multicolumn{2}{c}{Action} & \multicolumn{2}{c}{Emotion} \\ 
& CSDF & Accuracy & CSDF & Accuracy \\ \hline
ours & 0.91  & 0.82  & 0.73   & 0.50     \\ \hline
Aberman et al.\ \cite{aberman2020}  &  0.84 &  0.68  & 0.5   & 0.16    \\ \hline
Park et al.\ \cite{park2021diverse}  & 0.84  & 0.73  &  0.52  &  0.17    \\ \hline
\end{tabular}
\caption{Quantitative comparison on style transfer with the CMU Xia dataset.}
\label{table:com_st}
\end{table}

\textit{Quantitative evaluation-} We assessed the generation quality, style transfer, and content preservation using two metrics: Cosine Similarity of Deep Features (CSDF) \cite{zhang2018unreasonable} and recognition accuracy. Deep features were extracted using an emotion classifier and an action classifier trained on the CMU Xia dataset. The CSDF was computed between the real and generated motion samples based on the 256-dimensional feature vectors from the second-last fully connected layer. Higher CSDF values indicate better generation quality.

Recognition accuracy was evaluated by comparing the action and emotion labels of the generated sequences with those of the source and reference motions. A correct match in action labels signifies preserved content, while a match in emotion labels confirms successful style transfer. 

Table \ref{table:com_st} presents the quantitative comparison results of the proposed method, Aberman et al.\ \cite{aberman2020}, and Park et al.\ \cite{park2021diverse}. Our proposed D-LORD achieved high CSDF values and accuracy for both the action and emotion classifiers, outperforming the competing methods. These results indicate that the proposed method excels in emotion transfer and motion content preservation.

In summary, both qualitative and quantitative evaluations indicate that D-LORD outperforms state-of-the-art methods in generation quality, style transfer, and content preservation.

\subsubsection{Motion Retargeting}
Figure \ref{fig:com_mr} compares the motion retargeting results on the MHAD dataset using methods from Villegas et al.\ \cite{villegas2018neural}, D-LORD, and LORD with bone consistency loss. Motion retargeting involves transferring motion from one character to another with a different skeletal structure by adapting skeleton-independent features.

In this task, the class input is standardized as a T-pose across all identities, eliminating the need for double latent optimization (D-LORD). Hence, a single latent optimization approach (LORD) with bone consistency loss suffices. LORD captures identity as a class label and uses content embeddings to retain the specific motion features for reconstruction. As shown in Figure \ref{fig:com_mr}, LORD performs comparably to the Villegas et al.\ \cite{villegas2018neural} method.

\begin{figure}
  \centering
  \includegraphics[width=0.98\columnwidth]{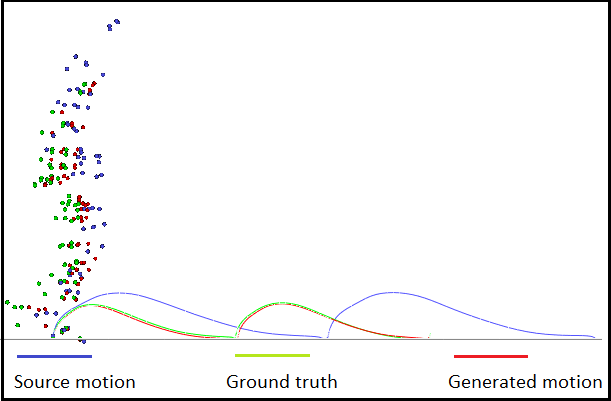}
   \caption{Left foot trajectory of source motion sequence, ground truth, and the generated motion sequence with D-LORD. D-LORD changes the skeleton as well as the gait cycle of the generated motion sequence.}
   \label{fig:gait}
\end{figure}

However, for more complex motion transfer tasks that require the adaptation of additional features, such as gait length, D-LORD becomes essential. For instance, Figure  \ref{fig:sample run} (third row) illustrates the transfer of motion between identities with different heights: identity $S_{11}$ (shorter) performing a step-up and identity $S_{101}$ (taller) performing a walk. Traditional retargeting methods would preserve the gait length of $S_{101}$ when transferring the walk motion to $S_{11}$. In contrast, D-LORD adjusts the gait length to match the shorter stride of $S_{11}$. Figure \ref{fig:gait} shows the trajectory analysis of the left foot. D-LORD's generated motion closely aligns with the ground truth gait cycle of $S_{11}$. In contrast, D-LORD adjusts the gait length to match the shorter stride of $S_{11}$. The trajectory analysis of the left foot in Figure \ref{fig:gait} shows that D-LORD's generated motion closely aligns with the ground truth gait cycle of $S_{11}$.

While LORD with bone consistency can handle standard motion retargeting, it struggles with transferring personalized motion features, such as a specific gait cycle. The "Root Motion Preservation" issue limits its style transfer capabilities when content characteristics (e.g., gait length) are part of the desired style. D-LORD addresses this limitation by transferring both the specified skeleton and content-specific features like gait length, enabling it to generate diverse motion sequences tailored to the specified skeleton.

Overall, both LORD and D-LORD effectively perform motion retargeting, adapting motions to varying skeletal structures. However, D-LORD's advanced disentanglement capabilities enable the transfer of additional features like gait. In contrast, methods like Villegas et al.\ \cite{villegas2018neural} only use a single input motion sequence and a T-pose, limiting their ability to perform full motion style transfer.

\section{Conclusions} \label{section:conclusions}
In this work, we introduced a novel data-driven, latent optimization-based disentanglement framework for motion stylization. The framework demonstrated superior performance compared to existing state-of-the-art techniques in motion style transfer. For motion retargeting, the framework employs LORD with bone consistency loss, yielding results comparable to other methods in the field. The primary objective of this research was to evaluate the applicability of the latent optimization-based disentanglement framework in various motion transfer applications. However, it should be noted that incorporating bone consistency loss alone does not guarantee an exact replication of bone lengths from the input skeleton, although visually similar skeletons are obtained. Additional skeleton constraints and post-processing of the output are required to achieve an exact copy of the input skeleton's bone lengths. This aspect will be further explored in future work. The effectiveness of the proposed framework was demonstrated using the CMU Xia dataset for motion style transfer, as well as the MHAD dataset and the RRIS ability dataset for motion retargeting.

\section*{Acknowledgment}
This research is supported by A*STAR under the National Robotics Programme (NRP) BAU grant, as part of the project titled "Assistive Robotics Programme" (Award No: M22NBK0074).

{\appendix
\subsubsection{Network architecture} 
The architecture of encoders and the decoder are shown in Table \ref{fig:ENA}.

\begin{table}
    \centering
\begin{tabular}{|c|c|} \hline  
    Encoders & Decoder  \\ [0.8mm] \hline 
  $CN_1K_7S_1P_3C_{64}, LRelu$   &  $FC_{Id*16}, LRelu$   \\ [0.8mm] 
   $CN_1SK_4S_2P_1C_{128}, LRelu$  & $FC_{Id*64}, LRelu$   \\ [0.8mm] 
    $CN_1K_4S_2P_1C_{256}, LRelu$  & $4*RB_(Id*64)$  \\ [0.8mm] 
    $CN_1K_4S_2P_1C_{256}, LRelu$  & $US(sf=2)$   \\ [0.8mm] 
     $CN_1K_4S_2P_1C_{256}, LRelu$  & $CN_1K_5S_1P_2C_{Id}, LRelu$    \\ [0.8mm]  
      $CN_1K_4S_2P_1C_{256}, LRelu$  & $UP(sf=2)$    \\ [0.8mm] 
      $FC_{256}, LRelu$ & $CN_1K_5S_1P_2C_{Id}, LRelu$   \\ [0.8mm]  
    $FC_{256}, LRelu$ & $CN_1K_7S_1P_3C_{Od}, tanh$  \\  [0.8mm] 
       $FC_{Ld}$ &     \\ [0.8mm] \hline 
\end{tabular}
\caption{Encoders and decoder architecture : $CN_I$-convId, $LRelu$-Leaky Relu, $K_i$-Kernal size $(i*i)$, $S_i$-Stride size $i$, $P_i$-Pedding size $i$, $C_i$-Output channel size $i$, $FC_i$-Fully connected layer output size $i$, $RB_i$-Residual block, output channel size $i$, $US(sf=i)$-upsampling layer with scale factor $i$, $Ld$-Latent dimension, $Id$-Input dimension, $Od$-Output dimension}
\label{fig:ENA}
\end{table}

The VAE model starts with a fully connected layer that reduces a 16-D input to an 8-D output, followed by two separate layers that output 2-D vectors for $\mu$ and $\sigma$. Using the reparameterization trick, a 2-D latent variable is sampled and then passed through two fully connected layers, sequentially expanding its size to 8-D and finally back to 16-D.

\subsubsection {Hyperparameters} 
The network hyperparameters for Stage 1 and Stage 2 were tuned using the RRIS validation dataset, testing various combinations of class, content, and aleatoric uncertainty (AU) embedding sizes ranging from 8 to 256 dimensions. Encoders trained on the RRIS dataset achieved the lowest reconstruction error with class and content embeddings of 128 dimensions and AU embeddings of 16 dimensions. As a result, the class and content embeddings were fixed at 128 dimensions ($c_{x_i}, e_{y_i} \in \mathbb{R}^{128}$), and the AU embeddings at 16 dimensions ($a_i \in \mathbb{R}^{16}$) for subsequent experiments. In Stage 1, the AU embeddings were regularized with additive Gaussian noise of random mean $\mu$ and fixed standard deviation $\sigma = 1$, using an $L_2$ regularization constant $\lambda$ of 0.001 (Equation 3). Latent optimization used the ADAM optimizer ($\beta_1 = 0.5$, $\beta_2 = 0.999$) for 500 epochs with a learning rate of 0.0001. Upon determining Stage 1 parameters, Stage 2 regularization coefficients were fine-tuned between 1 and 50, with values of 10 for all three coefficients ($\alpha_1$, $\alpha_2$, $\alpha_3$) achieving the lowest reconstruction error, and these were adopted for the remaining experiments (Equation \ref{eq_5}).


\begin{wrapfigure}{l}{1in}
\vspace{-0.5cm}
  \begin{center}
  \includegraphics[width=1in,keepaspectratio]{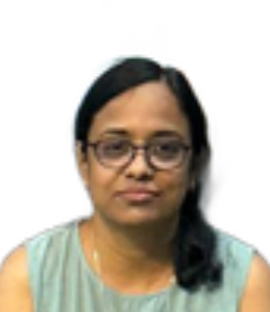} 
  \end{center}
\vspace{-0.5cm}
\end{wrapfigure}
\noindent \textbf{Meenakshi Gupta} received her B.E. degree from the University of Rajasthan, India. She completed her M.Tech. and Ph.D. from the Indian Institute of Technology Kanpur, India in 2006 and 2014, respectively. She was a postdoctoral fellow at the University of Sheffield, UK. Dr. Gupta has served as the Chief Technology Officer (CTO) at Invigilo Technology, Singapore, where she led initiatives in industrial video analytics. She also worked as a research fellow at Nanyang Technological University, Singapore, focusing on deep learning-based projects. Currently, she is a research fellow at the National University of Singapore, where she is involved in 5G-enabled multi-agent robotics digital-twin systems. Her research interests include artificial intelligence, deep learning, foundation models, computer vision, and robotics.

\vspace{0.25cm}

\begin{wrapfigure}{l}{1in}
\vspace{-0.5cm}
  \begin{center}
  \includegraphics[width=1in,keepaspectratio]{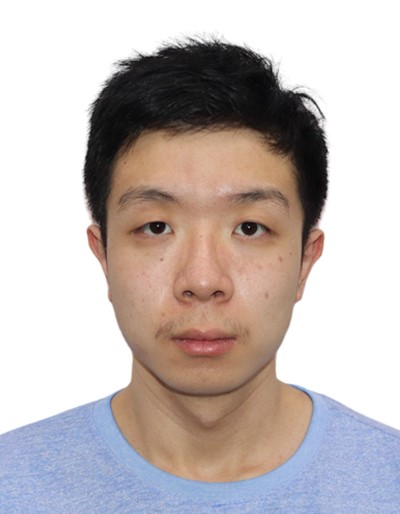} 
  \end{center}
\vspace{-0.5cm}
\end{wrapfigure}
\noindent \textbf{Mingyuan Lei} is a first-year PhD student at the College of Computing and Data Science (CCDS), Nanyang Technological University, Singapore. His research focuses on advancing AI-driven human-scene interaction within 3D environments. By incorporating language embeddings from LLMs and additional control mechanisms, Mingyuan aims to overcome limitations in conventional approaches to enhance the realism and adaptability of these interactions.

 \vspace{0.25cm}
\begin{wrapfigure}{l}{1in}
\vspace{-0.5cm}
  \begin{center}
  \includegraphics[width=1in,keepaspectratio]{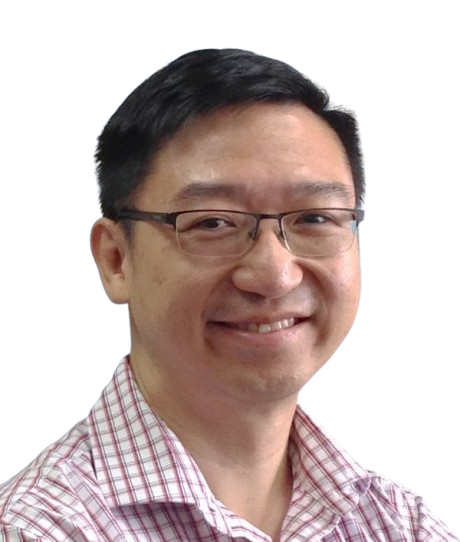} 
  \end{center}
\vspace{-0.5cm}
\end{wrapfigure}
\noindent \textbf{Tat-Jen Cham} is a Professor of Computer Science at the College of Computing and Data Science, Nanyang Technological University, Singapore. He received his BA in Engineering in 1993 and his PhD in 1996, both from the University of Cambridge. He was previously a Jesus College Research Fellow in Science (1996-97), and a research scientist at DEC/Compaq Research Labs in Cambridge, MA, USA (1998-2021). Tat-Jen has received multiple paper awards, including the best paper at ECCV'96, and is an inventor on eight patents. Tat-Jen has been the principal investigator on projects that include those based in the Rehabilitation Research Institute of Singapore (RRIS), the Singtel Cognitive \& AI Lab (SCALE@NTU), Singapore-ETH Centre’s Future Cities Lab, and the NRF BeingThere / BeingTogether Centres on 3D Telepresence. His current and past research services include being an Associate Editor for IEEE T-MM, CVIU, and IJCV, as well as an Area Chair for CVPR, ICCV, ECCV, and NeurIPS conferences. He was also a General Chair for ACCV 2014. Tat-Jen’s research interests are broadly in computer vision and machine learning, with a focus on deep learning generative methods that can exploit semantic and contextual cues, for applications such as 3D telepresence and metaverses.

\vspace{0.25cm}
\begin{wrapfigure}{l}{1in}
\vspace{-0.5cm}
  \begin{center}
  \includegraphics[width=1in,keepaspectratio]{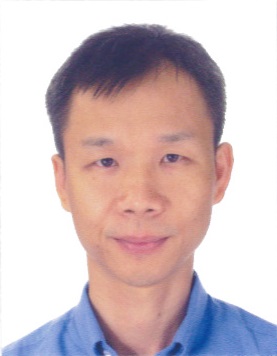} 
  \end{center}
\vspace{-0.5cm}
\end{wrapfigure}
\noindent \textbf{Dr. Lee Hwee Kuan} earned his Ph.D. from Carnegie Mellon University in Pittsburgh, PA, USA, in theoretical physics in 2001. Currently, he serves as the Head of the Imaging Informatics Division as well as the Deputy Director for Training and Talent Development at the Bioinformatics Institute, A*STAR in Singapore. Dr. Lee's expertise lies in the development and deployment of machine learning and deep learning algorithms. With a strong foundation in theoretical physics, Dr. Lee brings a unique perspective to the field of artificial intelligence, driving innovative solutions for both clinical and biological applications. 

At present, Dr. Lee Hwee Kuan leads a laboratory dedicated to the advancement of Artificial Intelligence (AI) research for clinical and biological purposes. His research endeavours encompass a broad spectrum of activities, from fundamental AI-centric investigations to practical AI applications. His laboratory addresses critical areas such as diagnostics in cancers, cardiology, dermatology, and interventional radiology. In the realm of biology, Dr. Lee's team develops bioinformatics pipelines for spatial omics and single-cell analysis, alongside pioneering AI applications in drug discovery. Beyond his role at the Bioinformatics Institute, Dr. Lee holds significant appointments in various local universities and research institutions.

\end{document}